\documentclass[10pt,twocolumn,letterpaper]{article}

\usepackage{iccv}              

\newcommand{\myparagraph}[1]{\smallskip\noindent\textbf{#1}}
\newcommand{\eat}[1]{}

\newcommand{\ours}{\texttt{SITE}\xspace}
\newcommand{\vlm}{{VLM}\xspace}
\newcommand{\si}{{SI}\xspace}
\newcommand{\vqa}{VQA\xspace}

\usepackage{soul}
\colorlet{darkred}{red!35}
\colorlet{lightred}{red!10}
\usepackage{float}

\newcommand{\mydarkredhighlight}[1]{%
  \begingroup
  \sethlcolor{darkred}%
  \hl{#1}%
  \endgroup
}

\newcommand{\mylightredhighlight}[1]{%
  \begingroup
  \sethlcolor{lightred}%
  \hl{#1}%
  \endgroup
}

\usepackage{pifont} 
\usepackage{booktabs} 
\usepackage[accsupp]{axessibility}  
\usepackage[T1]{fontenc}

\usepackage{listings} 
\definecolor{iccvblue}{rgb}{0.21,0.49,0.74}
\usepackage[pagebackref,breaklinks,colorlinks,allcolors=iccvblue]{hyperref}
\newcommand\blfootnote[1]
{\begingroup\renewcommand\thefootnote{}\footnote{#1}\addtocounter{footnote}{-1}\endgroup}

\def\paperID{15025} 
\def\confName{ICCV}
\def\confYear{2025}

\title{\ours: towards Spatial Intelligence Thorough Evaluation}

\author{Wenqi Wang\textsuperscript{1} \ \ \ Reuben Tan\textsuperscript{2} \ \ \ Pengyue Zhu\textsuperscript{1} 
\ \ \ Jianwei Yang\textsuperscript{2} \\   
Zhengyuan Yang\textsuperscript{2} \ \ \ Lijuan Wang\textsuperscript{2} \ \ \
Andrey Kolobov\textsuperscript{2} \ \ \ Jianfeng Gao\textsuperscript{2*} \ \ \ Boqing Gong\textsuperscript{1*}\\
$^{1}$\small{Boston University}, $^{2}$ Microsoft Research, Redmond \\
\href{https://wenqi-wang20.github.io/SITE-Bench.github.io/}{SITE-project-page}
\\
}

\begin{document}

\twocolumn[{%
\renewcommand\twocolumn[1][]{#1}%
\maketitle
\centering
  \centering
  \vspace{-10pt}
\includegraphics[width=0.98\linewidth]{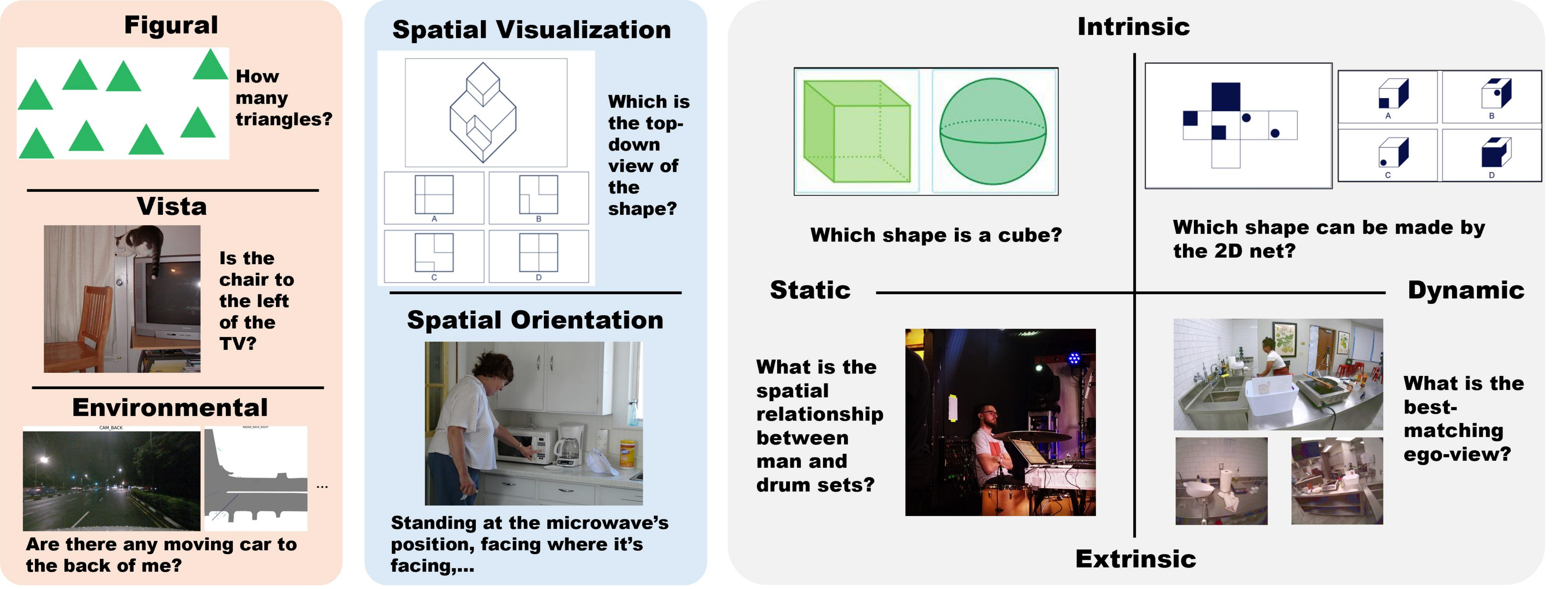}
\vspace{-5pt}
  \captionof{figure}{We introduce \ours, a comprehensive benchmark for evaluating large vision-language models' spatial intelligence (SI). Three SI classification systems drawn from cognitive science, corresponding to the three panels, drive the design of \ours.} 
  \vspace{20pt}
  \label{fig:three-systems}
}]

\blfootnote{$^*$ Equal advising}

\begin{abstract}
Spatial intelligence (\si) represents a cognitive ability encompassing the visualization, manipulation, and reasoning about spatial relationships, underpinning disciplines from neuroscience to robotics. 
We introduce \textbf{\ours}, a benchmark dataset towards \si Thorough Evaluation in a standardized format of multi-choice visual question-answering, designed to assess large vision-language models' SI across diverse visual modalities (single-image, multi-image, and video) and \si factors (figural to environmental scales, spatial visualization and orientation, intrinsic and extrinsic, static and dynamic). Our approach to curating the benchmark combines a bottom-up survey of existing datasets and a top-down strategy drawing upon three classification systems in cognitive science, which prompt us to design two novel types of tasks about view-taking and dynamic scenes.
Extensive experiments reveal that leading models fall behind human experts, especially in spatial orientation, a fundamental \si factor.
Moreover, we demonstrate a positive correlation between a model’s spatial reasoning proficiency and its performance on an embodied AI task. 
\end{abstract}    

\section{Introduction}
\label{sec:intro}
This work introduces \textbf{\ours}, a novel benchmark dataset towards \texttt{S}patial \texttt{I}ntelligence \texttt{T}horough \texttt{E}valuation, to assess the visuospatial ability of large vision-language models~(\vlm{s}). We achieve this by borrowing three classification systems about spatial intelligence~(\si) from the cognitive science literature~\cite{michael1957description, mcgee1979human, lohman1988spatial,carroll1993human,Kim_2021} to analyze vision-language tasks derived from 30 computer vision datasets. This process highlights a gap in existing benchmarks, leading us to design new tasks focusing on spatial orientation (view-taking) in static and dynamic contexts. We standardize all tasks to ease evaluation using a multiple-choice visual question answering (\vqa) format.

\si represents a cognitive capacity encompassing the visualization, manipulation, and reasoning about spatial relationships~\cite{gardner2011frames}. Figure~\ref{fig:three-systems} shows some SI tests at different scales (left panel), that require spatial visualization independent of one's viewpoint and spatial orientation due to changes of viewpoints (middle), extrinsic, and dynamic (right). SI is essential for many professions, including architecture, engineering, and the arts, and SI is visual --- some works in cognitive science use SI and ``visuospatial abilities'' interchangeably~\cite{shah2005cambridge}. While the computer vision community has addressed some components of \si through tasks such as object detection~\cite{Everingham10,lin2014microsoft,geiger2015kitti,yu2020bdd100k,xiang2016objectnet3d}, object referring~\cite{kazemzadeh2014referitgame,yu2016modeling,mao2016generation,yang2019fast}, 2.5D visual relationships~\cite{su20222}, and counting/localization in VQA~\cite{agrawal2016vqavisualquestionanswering,hudson2018gqa,marino2019ok}, progress has occurred mainly within individual domains or datasets. However, the advent of general-purpose large VLMs necessitates a unified testbed incorporating these diverse tasks and some uncaptured aspects of \si. This work presents such a benchmark and a systematic approach to building it.


The role of \si in AI models is highly tied to its role in human perception and reasoning. Kant argued that ``{space is the form of outer intuition}", asserting that our perception of the physical world is structured by spatial relations between objects~\cite{kant2003critique}. This concept is equally crucial for AI models, which have to develop SI to navigate and interact effectively in complex environments, especially with regard to tasks including but not limited to object manipulation~\cite{kim24openvla,team2024octo,ebert2021bridge,bharadhwaj2024roboagent} and navigation~\cite{Gu_2022,Anderson_2018_CVPR,shah2023lm,huang2023visual}. In this work, we specifically evaluate large \vlm{s}. These models have demonstrated impressive capabilities in visual reasoning and question answering, thus positioning them as key components for perception and reasoning in embodied AI agents and robotics~\cite{kim24openvla,zitkovich2023rt}. However, their ability to perform fine-grained spatial reasoning remains relatively limited and only partially assessed. We expect this work will facilitate a comprehensive and holistic evaluation of \vlm{s} across as broad a spectrum of \si as possible.

\begin{table*}
\small
    \centering
    \begin{tabular}{l|ccccc}
        \toprule
        & \ours (ours) & VSI-Bench~\cite{yang2024thinkingspacemultimodallarge} & 3DSRBench~\cite{ma20243dsrbenchcomprehensive3dspatial} & CVBench~\cite{tong2024cambrian1} & SpatialEval~\cite{wang2025picture}\\
        \midrule
        Input & natural/synthetic image, video & video & image & image & image\\
        Scale & figural, vista, environmental & environmental & vista & vista & figural, vista \\
        Spatial Visualization & \ding{51} & \ding{55} & \ding{55} & \ding{55} & \ding{55}\\
        Spatial Orientation & \ding{51} & \ding{51} & \ding{51} & \ding{55} & \ding{51}\\
        Dynamic & \ding{51} & \ding{51} & \ding{55} & \ding{55} & \ding{55}\\
        Intrinsic & \ding{51} & \ding{55} & \ding{55} & \ding{55} & \ding{51}\\
        \bottomrule
    \end{tabular}
    \vspace{-5pt}
    \caption{\ours \vs.\ similar efforts on benchmarking spatial intelligence. Besides being more diverse and comprehensive than existing datasets, \ours also introduces a structured classification system to better analyze spatial reasoning capabilities.}
    \label{tab:benchmark_comparison}
    \vspace{-10pt}
\end{table*}

The proposed \ours benchmark poses a stark contrast to several similar efforts about SI. Table~\ref{tab:benchmark_comparison} summarizes the key differences. The visual part in \ours combines natural images, synthetic images, multiple views, and videos, while the existing ones contain only natural images~\cite{tong2024cambrian1, ma20243dsrbenchcomprehensive3dspatial} or videos~\cite{yang2024thinkingspacemultimodallarge}. More importantly, these benchmarks measure limited aspects of spatial intelligence. For instance, CVBench~\cite{tong2024cambrian1} lacks viewpoint transformation. 3DSRBench~\cite{ma20243dsrbenchcomprehensive3dspatial} is limited to single-image questions, overlooking spatial reasoning in dynamic contexts. While VSI-Bench~\cite{yang2024thinkingspacemultimodallarge} uses videos to evaluate VLMs' capability to perform spatial reasoning across time, it is constrained to only indoor scenes. Finally, while counting and localization frequently appear in \vqa benchmarks~\cite{agrawal2016vqavisualquestionanswering,hudson2018gqa,liu2024mmbenchmultimodalmodelallaround}, tasks that require reasoning across multiple viewpoints remain largely unaddressed. These gaps manifest significant challenges in comprehensively evaluating VLMs' spatial intelligence.

To address the gaps, we approach the curation of \ours from two complementary paths: \textit{Bottom-Up} and \textit{Top-Down}. In the \textit{bottom-up} path, we survey 30 representative datasets and systematically extract vision-language tasks after careful filtering. The filtering comprises two phases. We first prompt large language models (LLMs) using the language part of the tasks to reduce costs, and we then filter the surviving tasks by jointly screening their vision-language modalities. Finally, we identify six core categories from the tasks. This bottom-up approach gives rise to 6,943 tasks, including 3,135 image-based QA pairs and 3,808 video-based QA pairs. The \textit{top-down} approach draws upon three classification systems of \si from the cognitive science literature, capturing \si's primary factors from different perspectives: scales (figural, vista, and environmental), view-taking (spatial visualization and orientation), intrinsic vs.\ extrinsic structures, and static vs.\ dynamic scenes. Investigation shows that the tasks resulting from the bottom-up path underrepresent the view-taking and dynamic factors, so we design two novel types of tasks using the Ego-Exo4D dataset~\cite{grauman2024egoexo4dunderstandingskilledhuman}, which is rich in camera views of dynamic events. \ours unions the bottom-up and top-down tasks and standardizes them for the ease of evaluations, with a total of 8,068 tasks, covering 30 existing benchmark datasets and 1 newly annotated dataset. 

Through this systematic approach, we provide a compact and comprehensive benchmark to analyze \vlm's spatial intelligence. We make the following key contributions.
\begin{enumerate}
    \renewcommand{\labelenumi}{\arabic{enumi})} %
    \item Comprehensive Spatial Intelligence Benchmark: We systematically analyze existing datasets and benchmarks, extracting and pooling relevant tasks towards a comprehensive \si evaluation dataset that covers a broad range of visuospatial reasoning tasks. 
    \item Cognitive Science Inspired Taxonomy and New Tasks: We refer to not one but three \si classification systems grounded in cognitive science when building our dataset. The ample references reveal a need for view-taking and dynamic tasks, following which we design two novel types of tasks to close the gap.
    \item Evaluation of Leading \vlm{s}. We use \ours, the resulting benchmark dataset, to extensively assess state-of-the-art  VLMs. Results show that existing VLMs especially struggle with spatial orientation tasks, falling significantly behind human performance.
    \item Evident Correlation Between \si and Embodied AI. Finally, we empirically demonstrate that \vlm{s} with high visuospatial ability also perform well on robot manipulation, with a correlation coefficient of 0.902. 
\end{enumerate}

\section{Related Work}
\label{sec:related}

\myparagraph{\si in cognitive science.} \si has historically been studied as the interaction of multiple sensory modalities, including vision, touch, and hearing~\cite{Eliot2002}. Out of these modalities, vision has been recognized as the dominant modality facilitating sensory integration~\cite{shah2005cambridge}, making visual-spatial ability the primary focus of mainstream spatial intelligence assessments. Psychologists and cognitive scientists have attempted to define and decompose spatial ability through factor-analytic studies~\cite{michael1957description, mcgee1979human, lohman1988spatial, carroll1993human, Uttal_Meadow_Tipton_Hand_Alden_Warren_Newcombe_2012, Hegarty_2004} based on numerous paper-and-pencil tests~\cite{eliot1983international}. Several core factors have consistently emerged in these studies, including Spatial Visualization, Spatial Relations, and Spatial Orientation~\cite{michael1957description, mcgee1979human, lohman1988spatial}. Later, Carroll et al.~\cite{carroll1993human} introduced an additional factor: Visual Memory. Uttal et al.~\cite{Uttal_Meadow_Tipton_Hand_Alden_Warren_Newcombe_2012} and Hegarty and Waller~\cite{Hegarty_2004} proposed new classification models for spatial intelligence. Due to variations in experimental paradigms and analytical methodologies, the definition of spatial ability remains highly inconsistent—a widely acknowledged consensus in cognitive science and psychology~\cite{uttal2013malleability}. In Section \ref{sec:grouping}, we will further elaborate on these concepts.

\myparagraph{\si in computer vision.} Spatial visual reasoning has long been an active research topic in computer vision. Initial efforts have mostly focused on constructing large-scale image-based datasets~\cite{agrawal2016vqavisualquestionanswering,hudson2018gqa,Liu2022VisualSR,johnson2016clevrdiagnosticdatasetcompositional}, that include spatial visual reasoning tasks, to evaluateVQA approaches. Notable examples such as  VQA~\cite{agrawal2016vqavisualquestionanswering}, GQA~\cite{hudson2018gqa}, VSR~\cite{Liu2022VisualSR}, and CLEVR~\cite{johnson2016clevrdiagnosticdatasetcompositional} incorporate spatial reasoning tasks, often evaluating a model’s ability to reason about spatial relationships between objects within an image~\cite{agrawal2016vqavisualquestionanswering}. However, these datasets generally focus on relatively simple and straightforward tasks, such as verifying the correctness of spatial relationships between objects in an image~\cite{agrawal2016vqavisualquestionanswering}. This limitation negatively affects their ability to evaluate more complex aspects of spatial intelligence. Meanwhile, advancements in 3D vision and autonomous driving have significantly enriched spatial visual task datasets. Pioneering datasets such as ScanNet~\cite{dai2017scannet} and NuScenes~\cite{qian2024nuscenesqamultimodalvisualquestion} provide high-quality 3D annotations, thus facilitating the construction of spatial reasoning benchmarks~\cite{chen2020scanrefer, azuma_2022_CVPR, qian2024nuscenesqamultimodalvisualquestion, omni3dbench}, which leverage multi-image and multi-view inputs to increase spatial task complexity. However, these benchmarks are still largely about static scenes. In contrast, our work on \ours aims to mitigate this limitation by providing a systematic evaluation of spatial intelligence at multiple scales and of both dynamic and static scenes.

\myparagraph{\si for benchmarking VLM models.} 
As mentioned earlier, many recent VLM benchmarks~\cite{fu2024mmecomprehensiveevaluationbenchmark,liu2024mmbenchmultimodalmodelallaround, li2024mvbenchcomprehensivemultimodalvideo, fu2024video,yu2024mm} have acknowledged the importance of spatial intelligence and included relevant evaluation questions. However, these benchmarks are often limited since spatial reasoning is generally treated as one among many tasks, scattered across other evaluation tasks of comprehension, reasoning, and perception, such as OCR and Math Reasoning. Existing image-based VLM benchmarks including MME~\cite{fu2024mmecomprehensiveevaluationbenchmark}, MMBench~\cite{liu2024mmbenchmultimodalmodelallaround}, SpatialEval~\cite{wang2025picture} and CVBench~\cite{tong2024cambrian1} comprise tasks including but not limited to object counting, localization, and question answering about spatial relationships. The Blink~\cite{fu2024blink} dataset is similar in nature to our \ours where it introduces evaluation tasks that involve spatial reasoning from multiple viewpoints and perspectives. Given the inherent difficulty in equipping VLMs with the capability to perform effective spatial reasoning, Cheng \etal~\cite{cheng2024spatialrgpt} propose a new data curation pipeline that leverages 3D scene annotations as well as a module for integrating depth information into VLMs. 
Similarly, several video understanding benchmarks for VLMs also include tasks related to spatial reasoning. Notable examples of such datasets include MLVU~\cite{MLVU}, MVBench~\cite{li2024mvbenchcomprehensivemultimodalvideo}, and VideoMME~\cite{fu2024video}. However, similar to the image counterparts, these video benchmarks do not systematically isolate spatial intelligence as a core focus of their evaluations. Additionally, recent works such as 3DSRBench~\cite{ma20243dsrbenchcomprehensive3dspatial} and SpatialEval~\cite{wang2025picture}, heavily emphasize their evaluations on different aspects of spatial reasoning but are primarily limited to single-image evaluations. Similarly, VSI-Bench~\cite{yang2024thinkingspacemultimodallarge} incorporates video-based spatial tasks but remains constrained to indoor environments. Despite the contributions of these works, the introduced benchmarks generally do not explicitly evaluate VLMs' spatial reasoning abilities in a structured and comprehensive manner while our \ours aims to bridge this gap by evaluating spatial intelligence across multiple aspects and diverse visual context.
\looseness=-1


\section{What Makes Spatial Intelligence (SI)?} \label{sec:grouping}

\begin{figure*}[t]
    \centering
    \includegraphics[width=\textwidth]{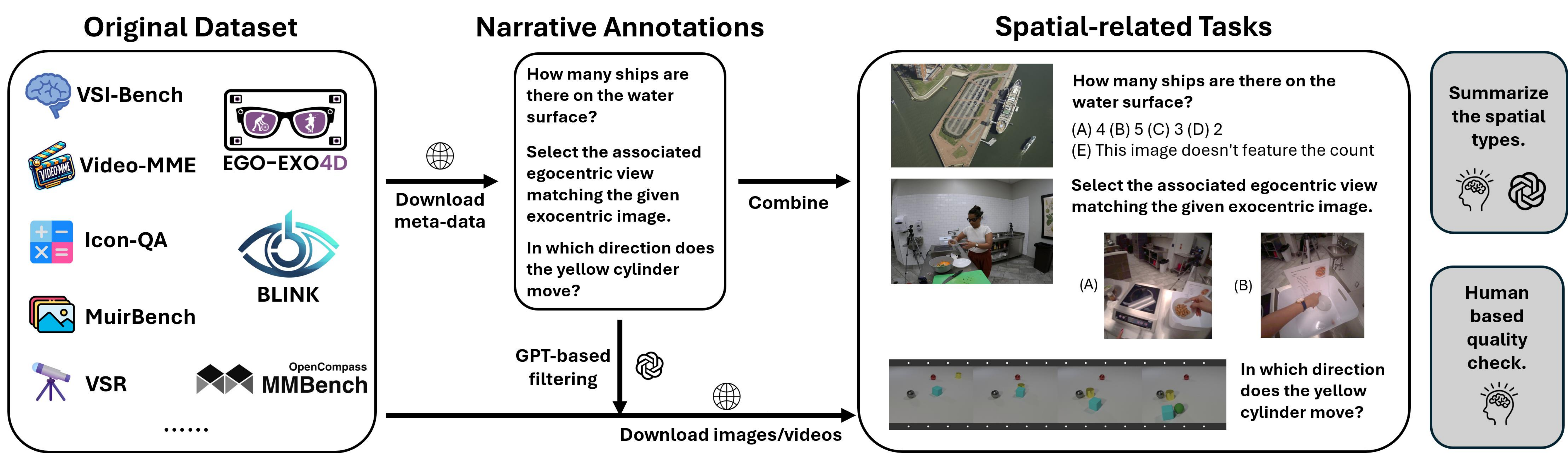}
    \vspace{-20pt}
    \caption{\textbf{Data collection pipeline for the \textit{bottom-up} part of our benchmark.} We conduct a large-scale effort to select image and video-based benchmarks that may contain SI tasks before using the GPT-4o model to filter out irrelevant evaluation samples. Finally, we generate 6 coarse categories and perform stratified sampling to obtain an even distribution over all SI categories. }
    \label{fig:data_collection_pipeline}
    \vspace{-10pt}
\end{figure*}

We aim to cover various factors of SI comprehensively, and yet the first challenge we encounter is the lack of a consensus on the categorization of SI~\cite{uttal2013malleability}. To the best of our understanding, at least three major classification systems about SI exist in cognitive science, and we refer to them all when building our benchmark.

\myparagraph{Figural, vista, and environmental SI.} Hegarty \etal study SI at figural, vista, and environmental scales~\cite{Hegarty_Montello_Richardson_Ishikawa_Lovelace_2006,montello1993scale} and note that ``processing spatial information at different scales of space involves different brain structures and mechanisms,'' inspiring us to include visual inputs at all these scales. Figural space is small relative to the human body and can be apprehended from a single viewpoint. The following cognitive tests are in this space: mentally folding paper, transforming shapes, and solving mazes. Vista space, such as single rooms and town squares, is larger than the body and remains apprehensible from a single viewpoint. Environmental space contains an individual, referring to cities, neighborhoods, \etc, which one must navigate to apprehend. Figure~\ref{fig:three-systems}(left) presents examples at the three scales.

\myparagraph{Spatial visualization \& orientation.} Hegarty and Waller compiled a summary of SI factors identified from primary cognitive studies~\cite{hegarty2005individual}, including spatial visualization (VZ), spatial orientation (SO), kinesthetic imagery~\cite{michael1957description}, relational reasoning~\cite{lohman1988spatial}, visual memory~\cite{carroll1993human}, and others. Of these, VZ and SO are considered fundamental (see Figure~\ref{fig:three-systems}(middle) for examples). Spatial visualization refers to the ability to mentally manipulate, rotate, or invert objects independently of one's own perspective, while spatial orientation involves imagining object appearances from differing observer viewpoints. It is important to note that these factors are derived from cognitive tests conducted prior to 2000, which primarily assess SI at figural and vista scales. It is also worth noting the extensive suite of cognitive tests about SI designed by Eliot and Smith~\cite{eliot1983international}.

\myparagraph{2x2 classification.} Uttal \etal categorize SI using a 2x2 classification system that relies on two distinctions, intrinsic vs.\ extrinsic and static vs.\ dynamic. Figure~\ref{fig:three-systems}(right) demonstrates this system. Intrinsic information refers to an object's defining features, parts, and the relationships among the parts. Extrinsic information, conversely, concerns the spatial relationships between objects within a group or their relation to a broader framework. Static tasks are about fixed spatial information (\eg, counting chairs in a dining room), while dynamic tasks involve movement and transformation (\eg, mental rotation of a 3D shape). 

\smallskip
In what follows, we employ all three categorization systems to drive our overarching design complementarily; meanwhile, we rely on them to balance and examine the coverage of the resulting benchmark from different aspects. Utilizing three rather than one categorization marks a significant difference between our work and most existing benchmarking efforts~\cite{yang2024thinkingspacemultimodallarge,ma20243dsrbenchcomprehensive3dspatial,liu2024mmbenchmultimodalmodelallaround,fu2024mmecomprehensiveevaluationbenchmark,lmms_eval2024,qian2024nuscenesqamultimodalvisualquestion,li2024mvbenchcomprehensivemultimodalvideo}.

\section{Data} \label{sec:data}
We adopt a two-stage approach to constructing \ours. To begin, in Section~\ref{sec: data_collection}, we compile a list of evaluation tasks from a suite of benchmarks that are focused on evaluating existing AI models' SI. Through a series of analysis and filtering steps, we construct a unified multiple-choice QA benchmark in a bid to standardize evaluations of SI in VLMs. Next, using the empirical analysis performed earlier as well as the classification framework introduced in Section \ref{sec:grouping}, we identify two underrepresented aspects of SI that are largely unaddressed by existing spatial reasoning benchmarks. To address these gaps, we propose new evaluation tasks to enable a more comprehensive and holistic evaluation of VLMs' spatial reasoning capabilities in Section~\ref{sec:proposed_tasks}. Finally, we describe the curation of our final \ours dataset in Section~\ref{sec:final_curation}.  

\subsection{Data Collection}\label{sec: data_collection}
Given the large number of isolated and fragmented evaluation benchmarks that focus on different aspects of SI, collecting a large-scale and comprehensive evaluation benchmark presents a significant challenge. With this challenge in mind, we propose a systematic data collection pipeline designed to filter and sample tasks relevant to SI in a structured manner, as Figure \ref{fig:data_collection_pipeline}.

\myparagraph{Data collection and filtering.} To begin, we manually select a set of representative VQA datasets that potentially cover some aspects of \si (Figure~\ref{fig:data_collection_pipeline}). The initial collection of datasets comprises 22 image datasets and 8 video datasets, where examples of the former include VSR~\cite{Liu2022VisualSR} as well as CV-Bench~\cite{tong2024cambrian1}, and the latter include VSI-Bench~\cite{yang2024thinkingspacemultimodallarge} and VideoMME~\cite{fu2024video}, respectively. We provide a detailed description of our selected benchmarks in the supplemental material. We note that we focus exclusively on the validation and test splits for each dataset. While the aforementioned datasets are generally large, they often only contain a subset of evaluation samples that are related to SI. Thus, it is necessary for us to filter out evaluation samples that are not relevant to the goal of evaluating spatial reasoning in VLMs. To filter out irrelevant evaluation samples, we adopt different strategies based on the existing annotations in each dataset.  For datasets that contain category labels for the evaluation samples, we only retain evaluation samples that fall under categories relevant to spatial reasoning while discarding the rest. For datasets without predefined labels, we use a pretrained Large Language Model (LLM) to perform filtering of their constituent evaluation samples as described below. 


\myparagraph{LLM filtering.} To further refine the quality and relevancy of the collected evaluation samples, we employ a filtering process by leveraging  GPT-4o~\cite{openai2024gpt4technicalreport}, a powerful LLM, to classify the evaluation samples (Figure~\ref{fig:data_collection_pipeline}). A major issue we faced in curating a comprehensive spatial reasoning benchmark is that different datasets use different labels for their evaluation samples, even though they might be evaluating similar aspects of SI. We tackle this issue by prompting the LLM to generate 6 coarse categories of tasks pertaining to SI. Specifically, we amass a set of original question labels from existing datasets and create a prompt for the LLM along with one to two sample datapoints as context for generating the task categories. The six coarse-level spatial intelligence categories are as follows: Counting and Existence (\textbf{Count.}), Spatial Relationship Reasoning 
(\textbf{Rel.}), Object Localization and Positioning (\textbf{Loc.}), 3D Information Understanding (\textbf{3D Inf.}), Multi-View Reasoning (\textbf{MultiV.}), and Movement Prediction and Navigation (\textbf{Mov.}). We proceed by classifying the valid spatial intelligence evaluation samples across all datasets into the abovementioned six categories.
\looseness=-1

Despite this categorization, many evaluation samples lack task-type labels or have noisy annotations. To refine the classification, we conduct a filtering stage using GPT-4o. We design a prompt template by incorporating key textual dataset information from each evaluation sample (e.g., questions, answers, options, and descriptions) along with carefully selected example data. The LLM is then queried to determine: \begin{enumerate*}[(1)] \item Whether the task pertains to spatial intelligence; \item If so, which coarse category it falls under. \end{enumerate*} We provide an example of the prompt template in the supplemental material.

\begin{table}[t]
    \centering
    \small
    \renewcommand{\arraystretch}{1.0} 
    \setlength{\tabcolsep}{6pt} 
    \label{tab:mathvista-stats}
    \begin{tabular}{l r}
        \toprule
        \textbf{Statistic} & \textbf{Number} \\
        \midrule
        Total questions & 8,068 \\
        \quad - 4-choice questions & 5,019 (62.2\%) \\
        \quad - 2-choice questions & 1,573 (19.5\%) \\
        \quad - 3/5/6-choice questions & 1,476 (18.3\%) \\
        \midrule
        \quad - Questions with annotations & 6,943 (86.1\%) \\
        \quad - Questions newly annotated & 1,125 (13.9\%) \\
        \midrule
        Source datasets & 31 \\
        \quad - Existing image datasets & 22 \\
        \quad - Existing video datasets & 8 \\
        \quad - Our newly annotated datasets & 1 \\
        \midrule
        Number of images & 13,172 \\
        Number of videos & 3,808 \\
        \bottomrule
    \end{tabular}
    \vspace{-5pt}
    \caption{\textbf{\ours benchmark statistics.}} \label{tab:data_comp}
    \vspace{-15pt}
\end{table}

\myparagraph{Statistics and Reform QA types.} After undergoing LLM-based filtering, the resulting dataset is reduced to 223,083 task examples, comprising 206,887 image-based and 16,196 video-based QA pairs relevant to spatial intelligence. We conduct a statistical analysis of the data across the coarse-level spatial categories and observe a significant data imbalance. Specifically, Relationship Reasoning (\textbf{Rel.}) and Counting and Existence (\textbf{Count.}) problems dominate the evaluation. In contrast, tasks such as Multi-View Reasoning (\textbf{MultiV.}) are underrepresented to a large degree. Additionally, due to the diverse data sources, the QA formats can vary considerably across different datasets. To ensure consistency and ease of evaluation, we standardize all tasks to a multiple-choice QA format, where we reformulate open-ended QA tasks accordingly.

\subsection{Novel Proposed Tasks} \label{sec:proposed_tasks}
To apply the cognitive science-based classification framework introduced in Section~\ref{sec:grouping}, we also conduct a fine-grained manual annotation of the filtered SI tasks. Following the classification workflow illustrated in Figure~\ref{fig:data_collection_pipeline}, we obtained the distribution of the collected dataset under different classification systems, as provided in the appendix. Our analysis reveals that there is a significant lack of tasks involving perspective transformations, which indicates that most existing SI evaluation tasks are constrained to reasoning from a fixed camera viewpoint. However, perspective transformation is especially essential for spatial reasoning in real-world scenarios such as navigation and route planning, where humans are naturally able to interpret spatial relationships from multiple viewpoints. To bridge this gap, we introduce two novel tasks specifically designed to evaluate \textbf{extrinsic-static} and \textbf{extrinsic-dynamic} spatial reasoning under perspective-transformed conditions.


\begin{figure*}
\vspace{-5pt}
    \centering
    \begin{minipage}{0.95\linewidth}
        \centering  
        \includegraphics[width=\linewidth]{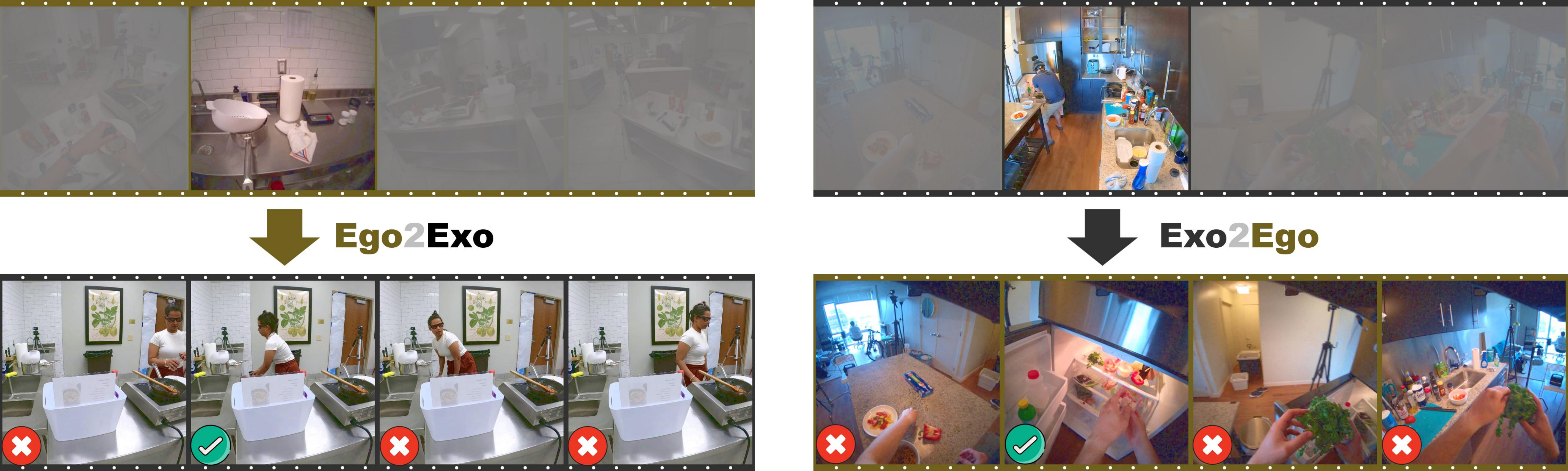}
        \vspace{-20pt}
        \caption{\textbf{Ego-Exo view association tasks.} The goal of this task is to pick the correct exocentric view given the egocentric view of a visual scene or vice versa.}
        \label{fig:view-association-tasks}
    \end{minipage}
    \vspace{-10pt}
\end{figure*}

\myparagraph{Data Preparation.} To assess models' perspective transformation abilities in real-world scenarios, we leverage videos and annotations from the  Ego-Exo4D~\cite{grauman2024egoexo4dunderstandingskilledhuman} dataset. Ego-Exo4D provides 5,035 takes across eight real-world scenarios, where each take includes at least one egocentric view and multiple exocentric views captured synchronously. This multi-view setting provides a strong foundation for our task design, enabling a robust evaluation of spatial intelligence across diverse viewpoints.

\begin{table}
    \centering
    \resizebox{\linewidth}{!}{ %
    \begin{tabular}{lcccc}
        \toprule
         & \multicolumn{2}{c}{\textbf{View Association}} & \multicolumn{2}{c}{\textbf{Frames Reordering}} \\
        \cmidrule(lr){2-3} \cmidrule(lr){4-5}
        Model & ego2exo & exo2ego & ego2exo & exo2ego \\
        \midrule
        \multicolumn{5}{l}{\textbf{Baseline \& Upperbound} on small subuset} \\
        Random & 0.0 & 0.0 & 0.0 & 0.0 \\
        Human performance & 100 & 100 & 98 & 96 \\
        InternVL-2.5-8B & 11.11 & -5.56 & -6.67 & -6.67 \\
        GPT-4o & 28.89 & 37.77 & 20.00 & 11.11 \\
        \midrule
        \multicolumn{5}{l}{\textbf{Open-source}} \\
        LLAVA-OneVision-0.5B & 0.46 & 3.38 & 3.61 & 3.85 \\
        LLAVA-OneVision-7B & \cellcolor{red!10} 21.80 & \cellcolor{red!10}10.10 & 2.01 & -4.41 \\
        Phi-3.5-Vision-4B & 5.09 & 2.11 & 4.42 & -0.28 \\
        InternVL-2.5-4B & 1.85 & -2.11 & \cellcolor{red!35}8.43 & -4.79 \\
        InternVL-2.5-8B & -5.56 & 5.91 & 5.22 & -0.66 \\
        QWen2.5-VL-3B & -0.93 & 0.84 & 3.61 & -2.54 \\
        QWen2.5-VL-7B & 5.09 & -3.80 & 7.63 & \cellcolor{red!35}4.23 \\
        \midrule
        \multicolumn{5}{l}{\textbf{Proprietary}} \\
        Gemini-1.5-pro & 15.30 & -1.27 & 6.83 & -4.04 \\
        GPT-4o & \cellcolor{red!35}35.70 & \cellcolor{red!35}20.70 & -2.01 & -5.16 \\
        \bottomrule
    \end{tabular}
    }
    
    \caption{\textbf{Evaluation on our proposed View Association and Frames Reordering tasks.} \mydarkredhighlight{Dark red}: Best among all models, \mylightredhighlight{light red}: Best  among open-source models.} 
    \label{tab:proposed_tasks_results}
    \vspace{-10pt}
\end{table}

\myparagraph{Ego-exo View Association.} As Figure~\ref{fig:view-association-tasks} shows, this extrinsic-static task requires VLMs to associate egocentric and exocentric views of the same visual scene. Given an image with an exocentric viewpoint, a model must select the best-matching egocentric image from a set of candidates. Conversely, if an egocentric image is given, the model has to select the best-matching exocentric image. To construct this task, we utilize fine-grained key step annotations provided in the original dataset and sample multiple frames around each key step. Then, we rely on qualified human annotators to select challenging distractor frames. 

\myparagraph{Shuffled Frames Reordering.} In this extrinsic-dynamic task, a model has to infer the correct temporal order from multiple viewpoints. Specifically, given a video clip, we extract the start and end frames from the egocentric video as reference points. Within this segment, we also sample 4 frames that capture key motion events from the exocentric views before shuffling them randomly. Finally, the model must predict the correct sequence by reasoning about motion dynamics across space and time. This task is also evaluated in reverse, by requiring models to predict the correct temporal order of egocentric frames based on the provided exocentric views. To ensure interpretability and feasibility of our proposed frame-reordering task, we also incorporate another round of human annotations to remove ambiguous and extremely difficult cases.


\subsection{\ours} \label{sec:final_curation}
To ensure a balanced representation of various \si factors, we perform stratified sampling on the collected data from Section~\ref{sec: data_collection} to achieve an even distribution across different spatial categories while maintaining diversity. Including the two newly proposed tasks, our final benchmark consists of 8,068 QA pairs, including 4,260 image-based QA pairs and 3,808 video-based QA pairs, covering 30 existing benchmark datasets and the Ego-Exo4D dataset. Table~\ref{tab:data_comp} shows some statistics.


\section{Experiments} \label{sec:exp}

\subsection{Benchmark Evaluation}

\myparagraph{Evaluation Models.} We evaluate 9 state-of-the-art VLMs that accept both image and video inputs, covering a diverse range of model architectures and parameter scales. From open-sourced models, we select LLAVA-OneVision~\cite{li2024llavaonevisioneasyvisualtask}, InternVL-2.5~\cite{chen2024internvl}, Qwen2.5-VL~\cite{qwen2.5-VL}, and Phi-3.5V~\cite{abdin2024phi3technicalreporthighly}. For proprietary models, we evaluate GPT-4o~\cite{openai2024gpt4technicalreport} and Gemini 1.5~\cite{gemini_1_5}. To ensure standardized and reproducible evaluation, we utilize the lmms-eval~\cite{lmms_eval2024} framework for benchmarking all models.

\myparagraph{Evaluation Metrics.} To ensure consistent and reliable evaluation of VLM-generated responses, we employ an LLM as part of an automated evaluation pipeline. In this study, we use GPT-4o for assessing model outputs. Since our benchmark follows a multiple-choice QA format with options of different lengths, the chance probability of guessing the correct options varies across questions. To mitigate this bias, we adopt a \textit{\textbf{Chance-Adjusted Accuracy}} (\begin{math}\mathcal{CAA}\end{math}) metric, which adjusts accuracy scores by accounting for the probability of random guesses, providing a more accurate measure of the model’s true reasoning ability beyond chance:
\vspace{-1.0em}
\begin{align}
\label{eq:adjusted_accuracy}
\mathrm{\mathcal{CAA}} 
= \left({\sum_{i=1}^{N} X_i \;-\; \sum_{i=1}^{N} \frac{1}{n_i}}\right)/
       \left({N \;-\; \sum_{i=1}^{N} \frac{1}{n_i}}\right)
\tag{1}
\end{align}
where $N$, $n_i$, and $X_i$ are the total number of questions, number of answer choices for the $i$-th multiple-choice question, and an indicator variable, respectively. If the model correctly answers the $i$-th question, we set $X_i =1$. Otherwise, we set $X_i = 0$. \begin{math}\mathcal{CAA}=1\end{math} when all predictions are correct ($X_i = 1$ for all $i$), indicating perfect performance; \begin{math}\mathcal{CAA}=0\end{math} when the preditions perform no better than random guessing ($\sum X_i = 
\sum_{i=1}^{N} \frac{1}{n_i}$); \begin{math}\mathcal{CAA}<0\end{math} reflects performace worse than random guess. This adjustment ensures that model performance is evaluated as the improvement beyond random chance, providing a fairer comparison across questions with different numbers of answer options.

\myparagraph{Baseline and Upper Bound.} We use random chance as the baseline performance, where the expected score is \textbf{0}, following the \begin{math}\mathcal{CAA}\end{math} metric in equation \ref{eq:adjusted_accuracy}. For the upper bound, we evaluate on six task categories, randomly sampling 50 QA pairs per task (see the \textbf{Tiny Subset} group in Table ~\ref{tab:full_csbench_results}). We then conduct human performance evaluation by selecting 7 human participants to complete the benchmark. \begin{math}\mathcal{CAA}\end{math} scores from all participants are computed and averaged to derive the final upper bound performance.

\subsection{Results on \ours}

\begin{table}
    \centering
        \resizebox{\linewidth}{!}{%
            \begin{tabular}{l|c|cccccc}
                \toprule
                \textbf{Model} & \textbf{Overall} & \textbf{Count} & \textbf{Loc} & \textbf{3D Inf} & \textbf{MultiV} & \textbf{Rel} & \textbf{Mov}  \\
                \midrule
                Random            & 0.0 & 0.0 & 0.0 & 0.0 & 0.0 & 0.0 & 0.0  \\
                \midrule
                \multicolumn{8}{l}{\textbf{Tiny Subset}} \\
                Human            & 67.5 & 66.0 & 83.3 & 54.7 & 87.5 & 73.0 & 52.5  \\
                InternVL-2.5-8B          & 34.3 & 48.5 & 46.8 & 9.32 & 8.51 & 45.6 & 23.7 \\ 
                GPT-4o            & 35.6 & 42.4 & 51.2 & 11.0 & 17.8 & 42.7 & 19.5 \\
                \midrule
                \multicolumn{8}{l}{\textbf{Open-source}} \\
                LLAVA-OV-0.5B    & 18.4 & 28.0 & 32.3 & 5.67 & 3.77 & 30.6 & 4.70  \\
                LLAVA-OV-7B    & 30.2 & 51.8 & 38.5 & 22.4 & \cellcolor{red!10}9.40 & \cellcolor{red!35}55.3 & 9.18   \\
                Phi-3.5-Vision    & 21.8 & 33.2 & 34.0 & 11.7 & 3.33 & 32.8 & 11.7  \\
                InternVL-2.5-4B    & 29.4 & 47.9 & 32.9 & 11.4 & 3.94 & 47.2 & 22.9  \\
                InternVL-2.5-8B    & \cellcolor{red!10}32.8
                & 47.1 & 37.0 & \cellcolor{red!10}23.2 & 9.05 & 47.6 & \cellcolor{red!35}28.7  \\
                Qwen2.5-VL-3B       & 29.5 & 45.6 & 37.5 & 13.2 & 7.14 & 45.6 & 18.8  \\
                Qwen2.5-VL-7B       & 31.4 & \cellcolor{red!35}52.6 & \cellcolor{red!10}44.1 & 9.42 & 1.08 & 51.5 & 18.9  \\
                \midrule
                \multicolumn{8}{l}{\textbf{Proprietary}} \\
                Gemini-1.5-Pro   & 32.5 & 48.0 & 45.8 & 25.3 & 5.33 & 48.8 & 18.4 \\
                GPT-4o         & \cellcolor{red!35}37.8 & 44.6 & \cellcolor{red!35}56.0 & \cellcolor{red!35}26.9 & \cellcolor{red!35}22.0 & 54.6 & 18.4  \\
                \bottomrule
            \end{tabular}
        }
        \vspace{-5pt}
        \caption{\textbf{Performance comparison on the full \ours benchmark.} \mydarkredhighlight{Dark red}: Best, \mylightredhighlight{light red}: Best  among open-source models.} 
        \label{tab:full_csbench_results}
        \vspace{-15pt}
\end{table} 

\paragraph{View Association.} We begin by reporting the results of our evaluation on our proposed tasks of view association and frames reordering in Table~\ref{tab:proposed_tasks_results}. To benchmark the difficulty of these two tasks, we conduct a human study and the results demonstrate that humans are actually very adept at understanding and reasoning about 3D visual scenes from both egocentric and exocentric viewpoints. We randomly sampled 30 questions from each of the two views and found that human participants achieved perfect accuracy (100\%). However, even state-of-the-art proprietary and open-sourced models such as GPT-4o and InternVL-2.5-8B achieve very low performances compared to the human level. This trend is also corroborated by the accuracy obtained by the abovementioned models on the full split of the view association task where GPT-4o and Gemini-1.5-pro achieve an average of 28.20\% and 7.03\%, respectively. Interestingly, while VLMs such as LLaVA-OneVision and Qwen2.5-VL have been demonstrated to perform well on conventional image and video question answering benchmarks like MME and VideoMME, their perception capabilities do not translate well to reasoning about spatial relationships and information beyond just perception. For instance, we observe that highly capable VLMs such as LLAVA-OneVision-7B and QWen2.5-VL-7B achieve very low average \begin{math}\mathcal{CAA}\end{math} of 15.95\% and 0.65\% on this view association task. We hypothesize that the lack of training data involving viewpoint transformations in large vision-language datasets—where most reasoning tasks rely directly on the camera's perspective—is the primary cause. The severe data imbalance observed during our dataset collection further supports this hypothesis.
\looseness=-1

\myparagraph{Frames Reordering.} Furthermore, we also observe a similar performance trend on our proposed task of temporal frames reordering (Table~\ref{tab:proposed_tasks_results} right). Once again, the results show that humans are able to understand the temporal occurrence of events from different viewpoints, where they achieve close to perfect accuracy of 98\% and 96\% in the ego2exo and exo2ego directions, respectively. Interestingly, the GPT-4o model experiences a sharp drop in performance on this task, as compared to the task of view association. This might suggest that the GPT-4o model is not able to understand the mapping between different viewpoints of temporal events. Additionally, it is notable that the large-scale and proprietary models are severely underperforming open-sourced VLMs such as InternVL-2.5-8B and QWen2.5-VL-7B. In fact, the Qwen2.5-VL-7B model achieves the best average performance of 5.93\%. One possible reason underlying this result is that the Qwen2.5-VL model is trained on video grounding data, which helps the model to learn a more effective understanding of time and consequently, the temporal order of events in videos.
\looseness=-1


\myparagraph{Evaluation on the full \ours.} We report our evaluation of state-of-the-art open-sourced and proprietary models in Table~\ref{tab:full_csbench_results}. For more detailed analysis, we break down the results of the evaluation across the six coarse categories of spatial intelligence tasks, as discussed in Section~\ref{sec:grouping}. To begin, the performance achieved by various open-sourced and proprietary models is consistent with our observations in Table~\ref{tab:proposed_tasks_results}. There is a large performance gap between the accuracy obtained by humans and state-of-the-art VLMs, which suggests that simply scaling up the amount of supervised fine-tuning (SFT) and instruction following multimodal data for pretraining may be inadequate in helping these VLMs to acquire effective spatial intelligence. It is also notable that humans only achieve an overall \begin{math}\mathcal{CAA}\end{math} score of 67.5\%, which hints at the difficulty of our proposed \ours. Interestingly, humans perform significantly worse on counting(e.g., counting in a long video), 3D understanding(e.g., inferring the camera's transformation matrix), and movement prediction(e.g., navigating in a long video) as compared to the other three categories. This result might be due to humans' attention bottlenecks in tracking multiple objects~\cite{franconeri2013flexible}, explaining why counting moving objects and tracking spatial transformations across time is very challenging. 
\looseness=-1

However, the best-performing VLM GPT-4o still underperforms the human performance by $\sim$32\%. The Qwen2.5-VL-7B and InternVL-2.5-8B models lead in overall accuracy with 31.4\% and 32.8\%, respectively. Despite containing much fewer model parameters, these open-sourced models perform competitively with GPT-4o and Gemini-1.5-Pro. Interestingly, we observe that 
Qwen2.5-VL-7B performs the best among all open-sourced VLMs on localization. This might be due to the pretraining recipe of Qwen2.5-VL-7B which also includes image and video grounding tasks~\cite{qwen2.5-VL}. We also see that multi-view reasoning is especially challenging for VLMs in general, where the performance obtained by GPT-4o is lower than that of human performance by over 70\% on the tiny subset. On the full split, all of the state-of-the-art VLMs obtain \begin{math}\mathcal{CAA}\end{math} of less than 10\%. One possible reason for the low performance is that these VLMs are generally not trained with different viewpoints for the same image or video.
\looseness=-1

From these empirical results, we also observe the benefits of using larger models. As evidenced by the consistent performance gains obtained by LLaVA-OneVision-7B and Qwen2.5-VL-7B over their smaller counterparts, VLMs with a higher number of parameters generally are able to perform spatial reasoning of visual scenes and environments much more effectively. However, it is notable that 3D understanding scores are consistently low across all VLMs, with most models scoring below 15\%, indicating a persistent challenge in understanding depth(e.g., reasoning the depth relationships between objects) and three-dimensional spatial transformations(e.g., folding a 2D grid into a cube).
\looseness=-1

\subsection{Spatial Intelligence on Downstream Tasks}

\begin{table}[ht]
\centering
\vspace{-5pt}
\resizebox{\linewidth}{!}{%
\begin{tabular}{lccc}
\toprule
\textbf{Model} & L2 Dist $\downarrow$ & SR (\%) $\uparrow$ & CAA $\uparrow$ \\
\midrule
LLaVA-OneVision-0.5B & 0.268 $\pm$ 0.241 & 0.0 & 18.4 \\
LLaVA-OneVision-7B & 0.142 $\pm$ 0.172 & 0.0 & 30.2 \\
Qwen2.5-VL-3B & 0.139 $\pm$ 0.153 & 0.0 & 29.5 \\
Qwen2.5-VL-7B & \textbf{0.030 $\pm$ 0.040} & \textbf{38.0} & \textbf{31.4} \\
\bottomrule
\end{tabular}}
\vspace{-5pt}
\caption{\textbf{Correlation between SI and robotics manipulation on Libero Spatial.} The Pearson correlation coefficient between the negated mean L2 distance and CAA score is 0.902.}
\label{tab:libero_results}
\vspace{-10pt}
\end{table}

To understand why spatial intelligence is important, we conduct a toy experiment where we evaluate multiple VLMs of varying sizes on other real-world embodied tasks using the LIBERO-Spatial~\cite{liu2023libero} dataset. Our goal is to analyze the relationship between performance on spatial intelligence benchmarks and a model's capability to perform well in real-world tasks. Thus, we fine-tune and evaluate both variants of the LLaVA-OneVision and Qwen2.5-VL model variants under the few-shot setting. Specifically, we only use 40-160 trajectories from each task in the spatial suite to train each model, but we do not notice much difference. We use a constant learning rate of 2e-5 and fine-tune each model for 30 epochs. In Table~\ref{tab:libero_results}, we report the mean L2 distance between the final positions of the target object and robot arm effector across all episodes as well as the  overall success rate. The negated mean L2 distance and CAA scores on our \ours benchmark across all VLMs have a positive Pearson Correlation Coefficient of 0.902. Notably, we also observe that the Qwen2.5-VL-7B model achieves a success rate of 38\% while the others fail completely. These results suggest that pretraining data recipes and scale are both important for improving spatial intelligence in VLMs. We provide a more detailed correlation analysis in the appendix, comparing the impact of different benchmarks on embodied tasks. Importantly, these results indicate that AI agents have to possess a high degree of spatial intelligence to reason and interact effectively in the physical world.
\looseness=-1
\section{Conclusion} \label{sec:conclusion}
In this work, we introduce \ours, a comprehensive benchmark focused on evaluating VLMs' ability to perform visuospatial reasoning. We pull tasks from 30 existing datasets and then design two novel types of tasks for view-taking and dynamic scenarios. 
Evaluation on \ours demonstrates a huge gap between humans and state-of-the-art VLMs. Moreover, we empirically demonstrate the positive correlation between the performance of VLMs on  \ours and a robot manipulation task. 
\looseness=-1

{
    \small
    \bibliographystyle{ieeenat_fullname}
    \bibliography{spatial}
}

\newpage
\clearpage
\setcounter{page}{1}
\maketitlesupplementary
\setcounter{section}{0}
\renewcommand*{\thesection}{\Roman{section}}
\renewcommand*{\thesubsection}{\arabic{subsection}}
\renewcommand*{\thesubsubsection}{\roman{subsubsection}}


\section{Datasets Surveyed} \label{sec:dataset-collection}
As mentioned in the main text, we have collected a total of 31 computer vision datasets, comprising 22 image-based datasets, 8 video-based datasets and 1 newly annotated dataset.

\myparagraph{Image-based datasets.} We have collected the samples from the validation/test split of the following 22 image-based datasets: 
\begin{itemize}
    \item Blink~\cite{fu2024blink}: a comprehensive benchmark designed to evaluate multimodal large language models(MLLMs) across broad visual perception tasks.
    \item CLEVR~\cite{johnson2016clevrdiagnosticdatasetcompositional}: a visual question answering dataset containing various aspects of visual reasoning tasks.
    \item CVBench~\cite{tong2024cambrian1}: a vision-centric benchmark evaluating 2D and 3D understanding of the models. 
    \item GQA~\cite{hudson2018gqa}:  a visual question answering benchmark designed to test compositional reasoning and spatial understanding constructed from structured scene representations.
    \item IconQA~\cite{lu2022iconqanewbenchmarkabstract}: a dataset targeting diagram-based question answering, challenging models to interpret and reason over abstract visual representations.
    \item LogicVista~\cite{xiao2024logicvistamultimodalllmlogical}: a benchmark aimed at assessing the logical reasoning capabilities of MLLMs through structured visual tasks.
    \item MMBench~\cite{liu2024mmbenchmultimodalmodelallaround}: a large-scale benchmark for evaluating the performance of multimodal models across a wide range of vision-language tasks.
    \item MME~\cite{fu2024mmecomprehensiveevaluationbenchmark}: a comprehensive evaluation benchmark for MLLMs, covering various aspects of perception and cognition abilities.
    \item MME-RealWorld~\cite{zhang2024mme}: a real-world multimodal evaluation benchmark that tests models on practical perception tasks.
    \item MMIU~\cite{meng2024mmiu}: a comprehensive benchmark designed to evaluate multi-image tasks on MLLMs.
    \item MMTBench~\cite{mmtbench}: an evaluation benchmark containing massive multimodal tasks from various scenarios.
    \item MMVet~\cite{yu2024mm}: an evaluation suite focusing on the integration of multiple Vision-Language capabilities.
    \item MMVP~\cite{tong2024eyes}: an MLLM evaluation benchmark focusing on CLIP-blind pairs.
    \item MuirBench~\cite{wang2024muirbenchcomprehensivebenchmarkrobust}: a comprehensive benchmark targeting robust multi-image understanding abilities of multimodal models.
    \item SAT~\cite{ray2024satspatialaptitudetraining}: a spatial training dataset with static and dynamic spatial reasoning tasks.
    \item SeedBench~\cite{li2023seed,li2023seed2,li2024seed2plus}: an evaluation benchmark for generative comprehension capabilities of MLLMs.
    \item SpatialEval~\cite{wang2025picture}: a novel benchmark that covers different aspects of spatial reasoning in textual and visual formats.
    \item SPEC~\cite{peng2024synthesize}: a synthetic dataset designed to test the fine-grained vision-language understanding of models.
    \item VQAv2~\cite{lei2018tvqa}: an enhanced version of the Visual Question Answering dataset, providing more balanced question-answer pairs to reduce language biases and better evaluate visual understanding.
    \item VSR~\cite{Liu2022VisualSR}: a benchmark designed to assess visual spatial reasoning capabilities within images.
    \item VStarBench~\cite{vstar}: a visual question answering benchmark that focuses on detailed visual grounding on high-resolution images. 
    \item 3DSRBench~\cite{ma20243dsrbenchcomprehensive3dspatial}: a comprehensive 3D spatial reasoning benchmark on diverse entities.
\end{itemize}

\myparagraph{Video-based QA datasets.} Our video-based QA samples are collected from the following 8 video-based datasets: 
\begin{itemize}
    \item ActivityNetQA~\cite{yu2019activityqa}: a large-scale video question answering dataset based on ActivityNet, designed to evaluate models' abilities to comprehend and reason about complex human activities in videos.
    \item MLVU~\cite{MLVU}: a comprehensive benchmark designed for long video understanding 
    \item MVBench~\cite{li2024mvbenchcomprehensivemultimodalvideo}: a comprehensive benchmark for multimodal video understanding, assessing models on a variety of tasks.
    \item Open-EQA~\cite{OpenEQA2023}: an open-ended embodied question answering dataset that tests models' abilities to interact with and reason about 3D environments through videos and natural language queries.
    \item TGIFQA~\cite{jang2017tgifqaspatiotemporalreasoningvisual}: a dataset for spatiotemporal reasoning in video question answering through tasks like action recognition and repetition counting.
    \item TVQA~\cite{lei2018tvqa}: a video question answering dataset constructed from TV shows, focusing on temporal and contextual reasoning.
    \item VideoMME~\cite{fu2024video}: a comprehensive benchmark for evaluating multimodal large language models on video understanding tasks.
    \item VSI-Bench~\cite{yang2024thinkingspacemultimodallarge}: a benchmark designed to evaluate spatial reasoning abilities of MLLMs, focusing on tasks that require understanding spatial relationships within indoor scenes.
\end{itemize}

\myparagraph{Ego-Exo4D}~\cite{grauman2024egoexo4dunderstandingskilledhuman}: a large-scale, multimodal, multiview video dataset capturing skilled human activities from synchronized egocentric and exocentric perspectives. It encompasses over 1,286 hours of video data collected from 740 participants across 13 cities, featuring diverse tasks such as cooking, sports, and music. The dataset includes rich annotations like expert commentary, narrate-and-act descriptions, and atomic action labels, supporting benchmarks in fine-grained activity recognition, proficiency estimation, cross-view translation, and pose estimation.

\section{Category Filtering}
After we collected all the data and their text annotation information (questions, options, answers, descriptions, prompts), we used GPT-4o and manual inspection to derive 6 coarse level classifications of spatial intelligence for the data with category labels; for the data without category labels, we carefully designed the following prompts and used GPT-4o to perform few-shot classification of the text annotation information, shown as Figure~\ref{fig:prompt-1}, ~\ref{fig:prompt-2} and ~\ref{fig:prompt-3}. The six categories of spatial intelligence are defined and described as follows:
\begin{enumerate}
    \item \textbf{Counting \& Existence.} Evaluates the model's ability to detect and quantify object occurrences within static images or video sequences. This includes recognizing the presence or absence of specific objects and accurately counting their instances across frames.

    \item \textbf{Spatial Relationship Reasoning.} Assesses the model’s capacity to infer relative spatial relationships between objects. This encompasses understanding positional attributes such as proximity, occlusion, containment, and directional relations (e.g., left/right, above/below).
    
    \item \textbf{Multi-View Reasoning.} Measures the model’s ability to integrate and interpret information across multiple viewpoints. This includes understanding object appearances from different perspectives, reasoning about occluded or unseen parts, and reconstructing spatial arrangements from limited observations.
    
    \item \textbf{3D Information Understanding.} Evaluates the model’s capability to perceive and represent three-dimensional object properties. This involves recognizing shape, depth, surface structure, and spatial extent, as well as reasoning about object interactions in a 3D environment.
    
    \item \textbf{Object Localization \& Positioning.} Tests the model’s accuracy in determining object locations within an image or scene. This includes detecting precise spatial coordinates, generating bounding boxes or keypoints, and performing spatial alignment.
    
    \item \textbf{Movement Prediction \& Navigation.} Assesses the model’s ability to predict object motion and infer navigational paths within dynamic environments. This includes trajectory forecasting, motion intent recognition, and decision-making based on spatial and temporal cues.
\end{enumerate}

\section{Dataset Statistics}\label{sec:dataset-statistics}
Besides, we include more dataset statistics, a radar chart comparing various models, more examples in our dataset, an example about the frame-reordering task, and a bar chart to reveal data decomposition. 









\begin{figure}[t]
  \centering
  \begin{subfigure}[t]{0.91\linewidth}
    \includegraphics[width=\linewidth]{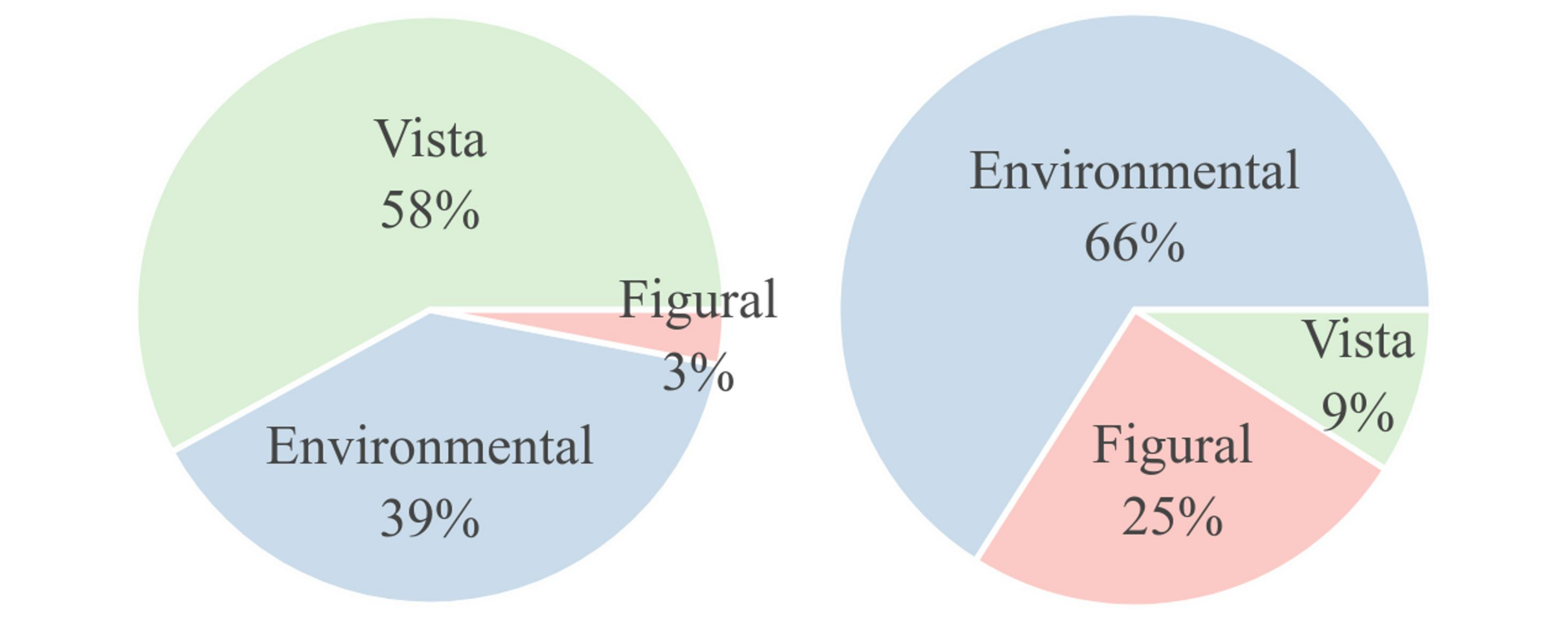}
    \caption{Figural, vista, and environmental SI.}
  \end{subfigure}
  \vspace{2mm}
  \begin{subfigure}[t]{0.91\linewidth}
    \includegraphics[width=\linewidth]{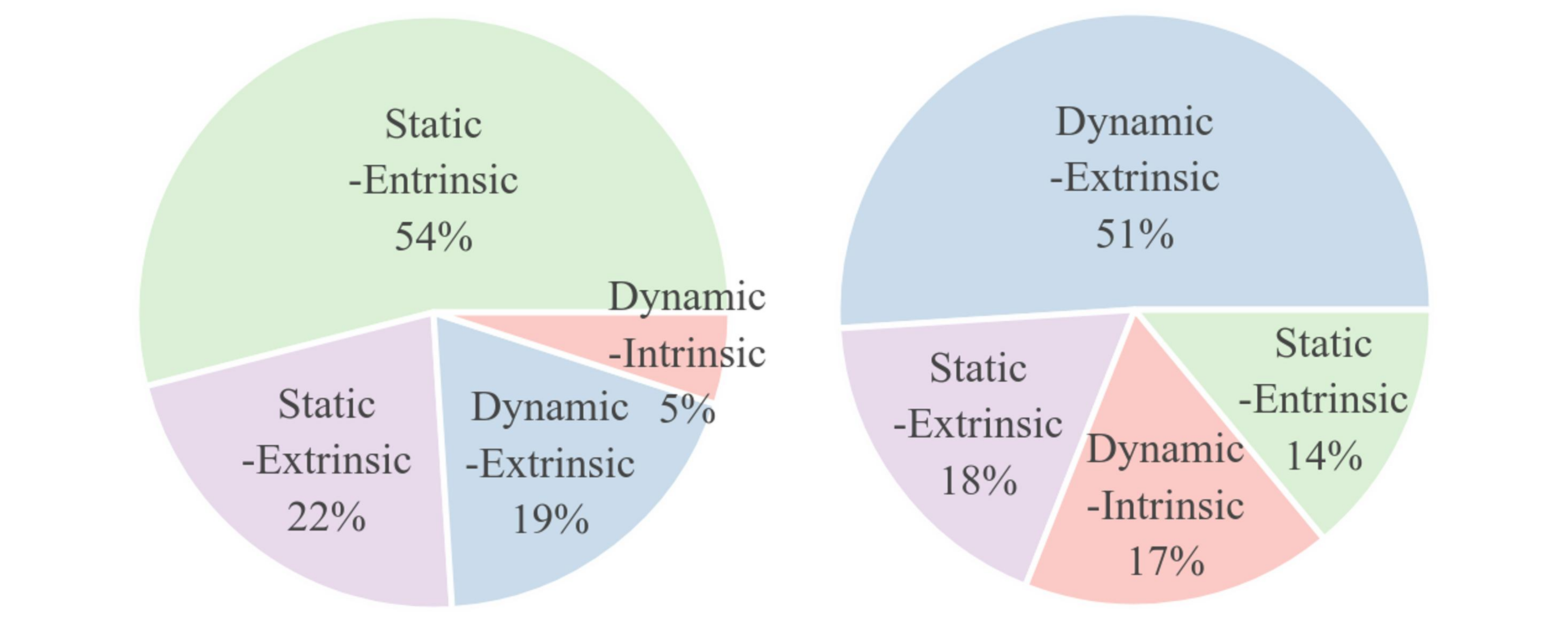}
    \caption{Spatial visualization \& orientation.}
  \end{subfigure}
  \vspace{2mm}
  \begin{subfigure}[t]{0.91\linewidth}
    \includegraphics[width=\linewidth]{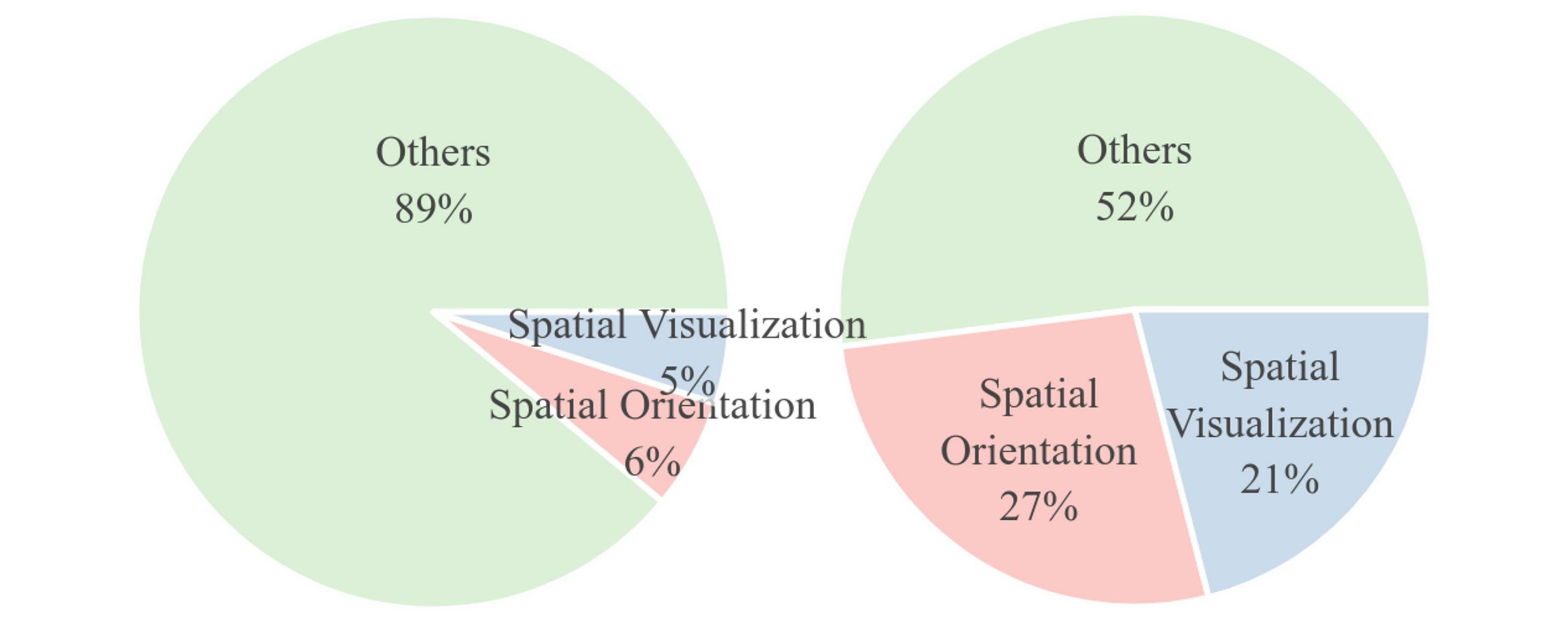}
    \caption{2x2 classification.}
  \end{subfigure}
  \caption{Category distribution by cognitive classification systems. \textbf{Left:} distribution before balancing. \textbf{Right:} our final benchmark's distribution.}
  \label{fig:right}
\end{figure}

\myparagraph{Correlation Analysis.} We collect performance scores from four vision-language models—Qwen2.5-VL and InternVL-2.5 series—across various benchmarks, along with their corresponding performances on LIBERO-Spatial when used as VLA backbones. We compute the Pearson correlation coefficient to measure the linear relationship between benchmark performance and LIBERO-Spatial results, as shown in Table~\ref{tab:libero_results_supp}. The analysis reveals that our SITE benchmark exhibits one of the strongest positive correlations, indicating that spatial intelligence plays a critical role in robotic manipulation tasks compared to general VQA capabilities (e.g., RealWorldQA, Q-Bench), OCR capabilities (OCRBench), scientific knowledge (ScienceQA), and object probing (POPE). Notably, MathVista also demonstrates a high positive correlation. We hypothesize that this may be attributed to the role of reasoning ability in enhancing performance on embodied tasks.


\begin{figure*}[htbp]
  \centering
  \begin{minipage}{0.95\linewidth}
    \centering
    \includegraphics[width=\linewidth]{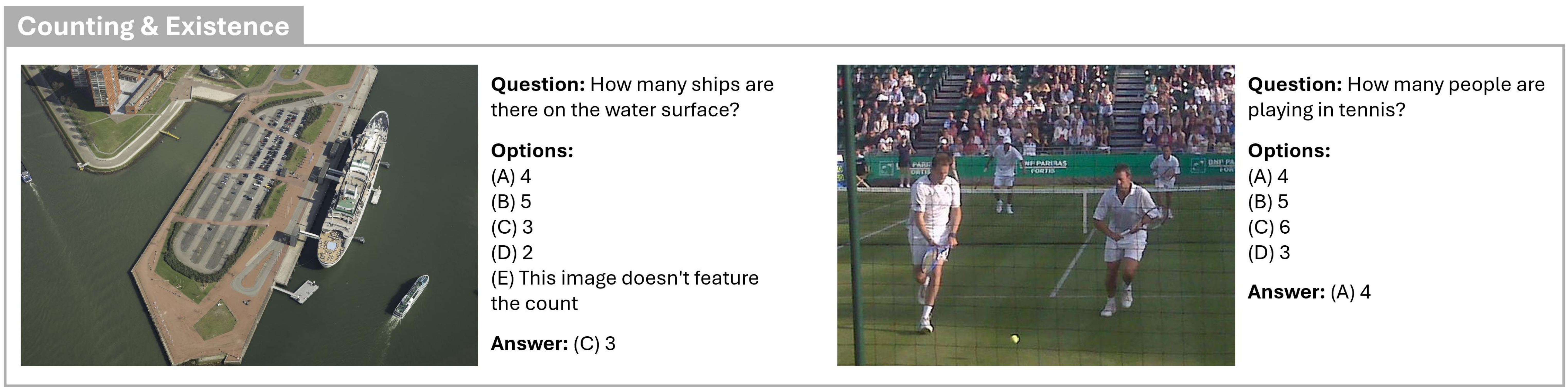}
  \end{minipage}\vspace{0.5em}

  \begin{minipage}{0.95\linewidth}
    \centering
    \includegraphics[width=\linewidth]{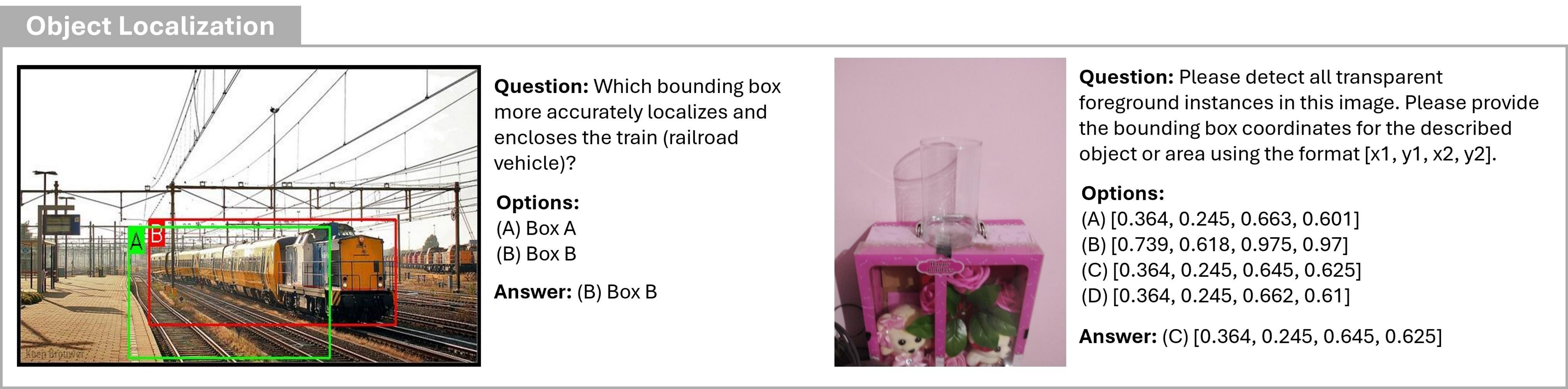}
  \end{minipage}\vspace{0.5em}

  \begin{minipage}{0.95\linewidth}
    \centering
    \includegraphics[width=\linewidth]{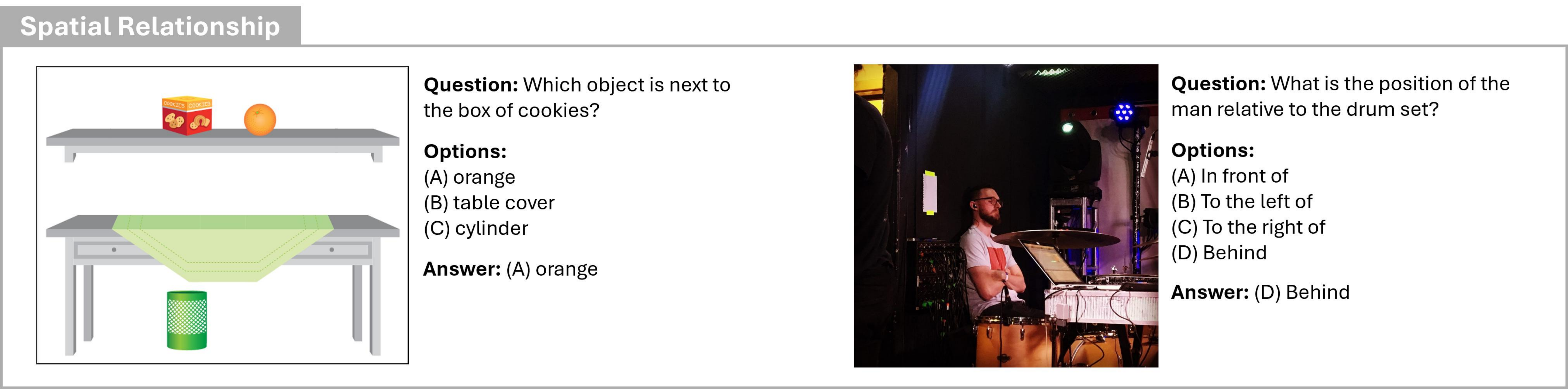}
  \end{minipage}\vspace{0.5em}

  \begin{minipage}{0.95\linewidth}
    \centering
    \includegraphics[width=\linewidth]{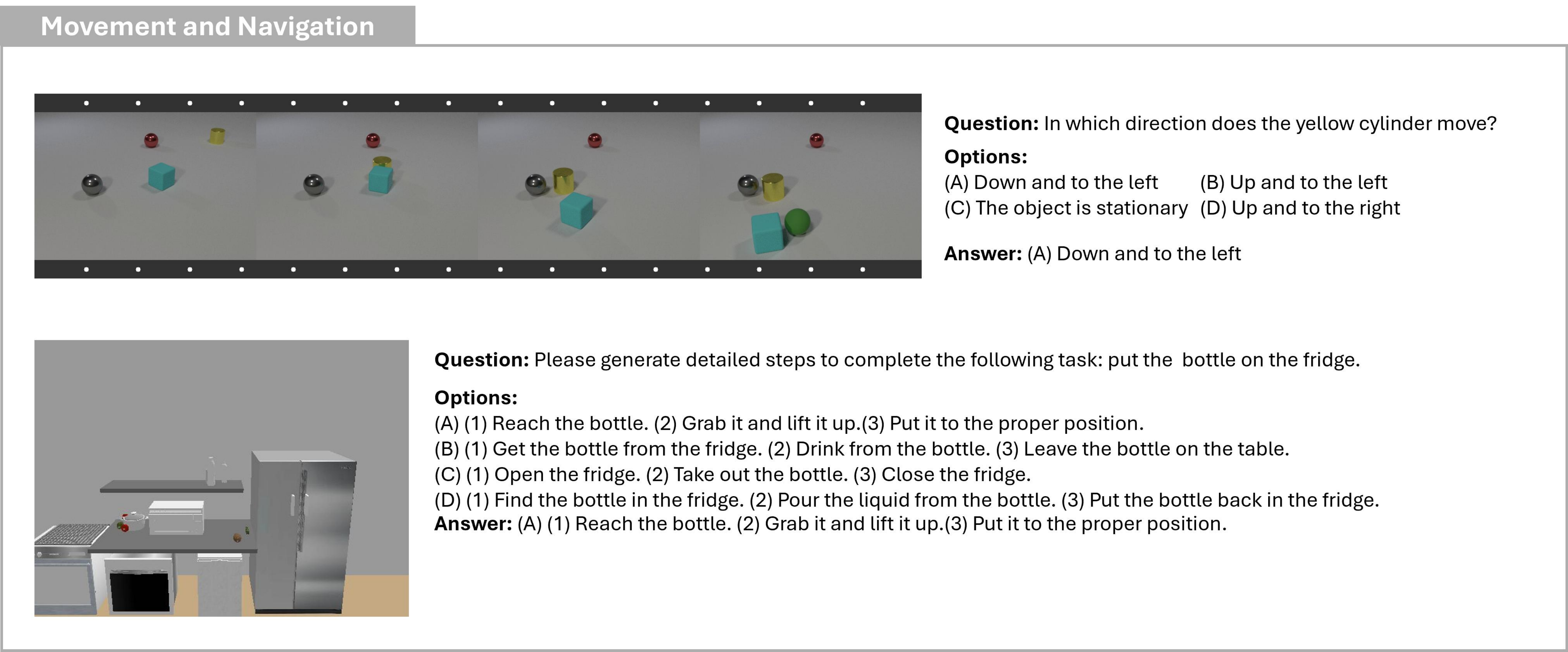}
  \end{minipage}
  \caption{Two samples for each spatial category. (Part I)}
  \label{fig:vertical4}
\end{figure*}

\begin{figure*}[htbp]
  \centering
  \begin{minipage}{0.95\linewidth}
    \centering
    \includegraphics[width=\linewidth]{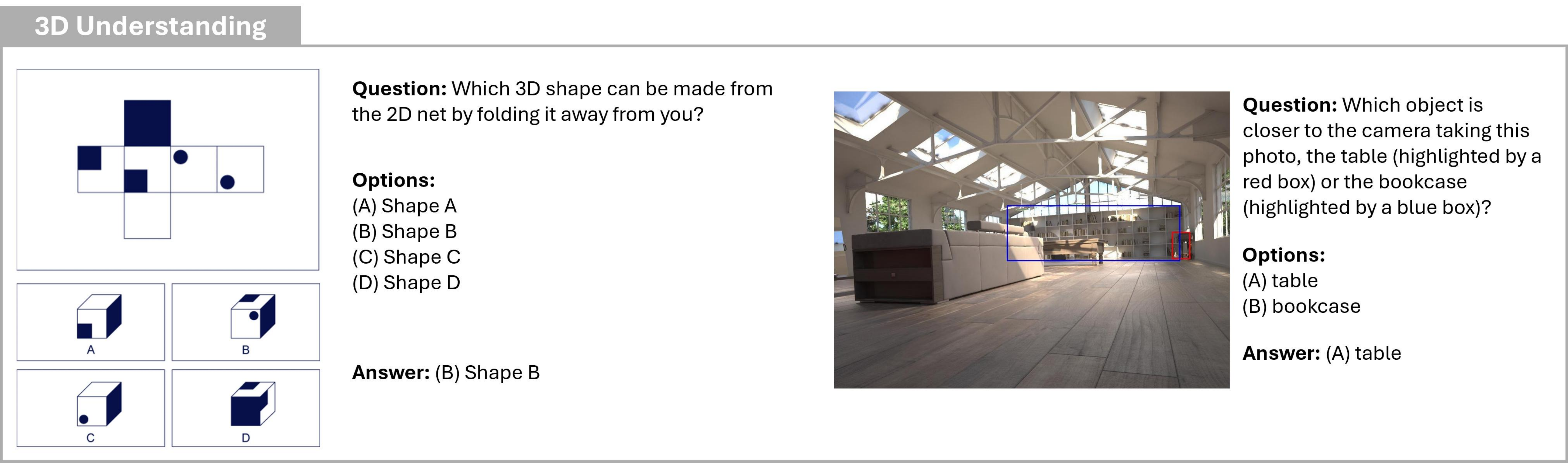}
  \end{minipage}\vspace{0.5em}

  \begin{minipage}{0.95\linewidth}
    \centering
    \includegraphics[width=\linewidth]{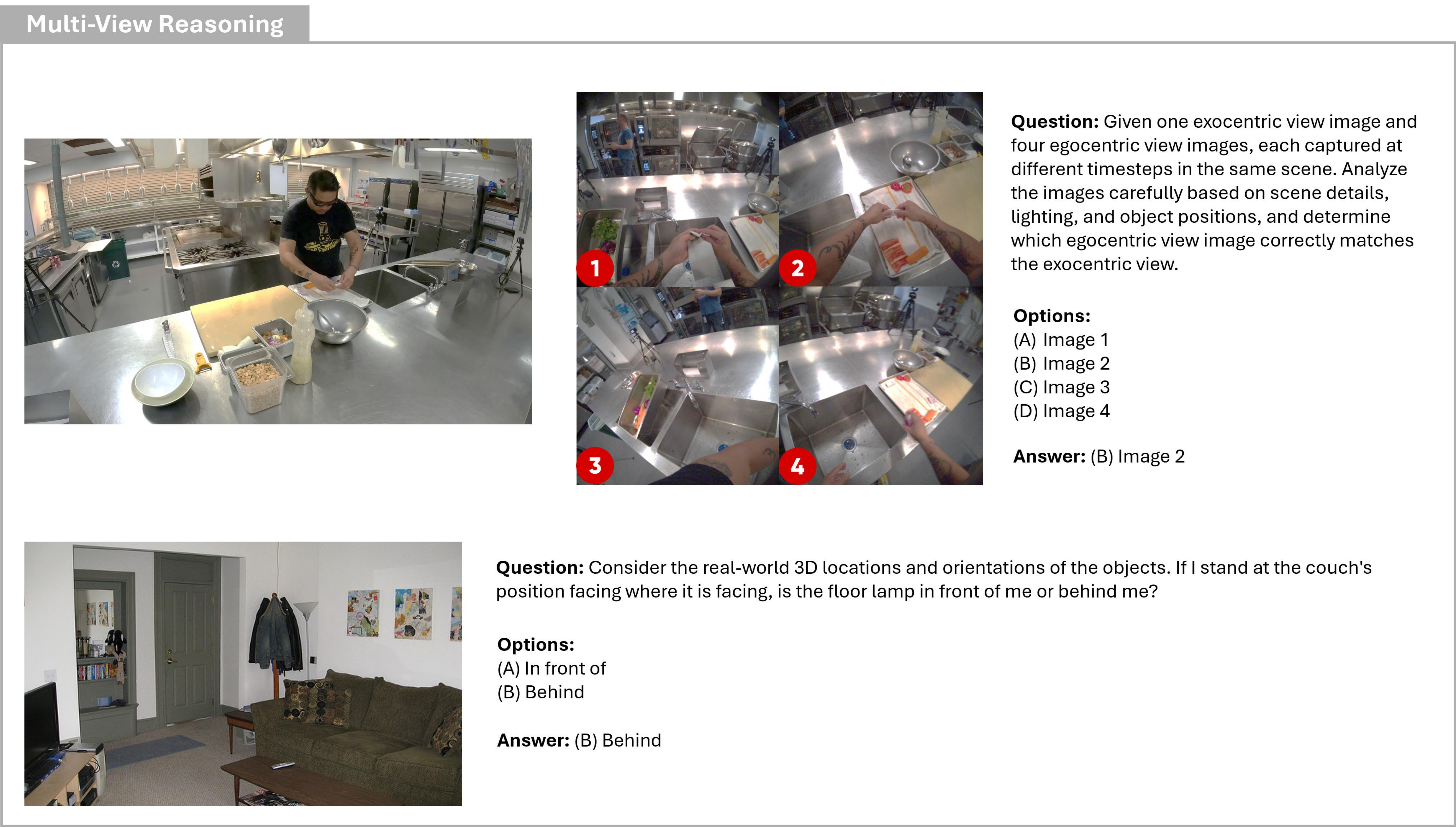}
  \end{minipage}\vspace{0.5em}

  \caption{Two samples for each spatial category. (Part II)}
  \label{fig:vertical4}
\end{figure*}

\begin{table*}
\centering
\vspace{-5pt}
\resizebox{\linewidth}{!}{%
\begin{tabular}{lccccccccc}
\toprule
\textbf{Model} & L2 Dist $\downarrow$ & Sim SR (\%) $\uparrow$ & MathVista $\uparrow$ & 
POPE $\uparrow$ & ScienceQA $\uparrow$ & OCRBench $\uparrow$ & 
RealWorldQA $\uparrow$ & QBench $\uparrow$ & \ours $\uparrow$ \\
\midrule
LLaVA-OV-0.5B & 0.268 $\pm$ 0.241 & 0.0 & 35.9 & 87.8 & 67.5 & 58.3 & 51.8 & 62.5 & 18.4 \\
LLaVA-OV-7B & 0.142 $\pm$ 0.172 & 0.0 & 62.6 & 88.4 & 95.4 & 62.2 & 69.9 & 78.9 &  30.2\\
Qwen2.5-VL-3B & 0.139 $\pm$ 0.153 & 0.0 & 61.2 & 85.9 & 81.4 & 82.8 & 65.5 & 74.9 & 29.5 \\
Qwen2.5-VL-7B & 0.030 $\pm$ 0.040 & 38.0 & 68.1 & 85.9 & 89.0 & 88.8 & 68.4 & 77.7 & 31.4 \\
Correlation & - & - & \textbf{0.935} & -0.602 &  0.749 &  0.832 &  0.842 & 0.847 & \textbf{0.902} \\
\bottomrule
\end{tabular}}
\vspace{-5pt}
\caption{\textbf{Correlation between SI and robotics manipulation on Libero Spatial.} The Correlation row shows the Pearson correlation coefficient between the negated mean L2 distance and different benchmark scores. The \textbf{bold} numbers show a higher correlation.}
\label{tab:libero_results_supp}
\vspace{-10pt}
\end{table*}


\begin{figure*}[t]
  \centering
  \begin{minipage}[c]{0.48\textwidth}
    \centering
    \includegraphics[width=\linewidth]{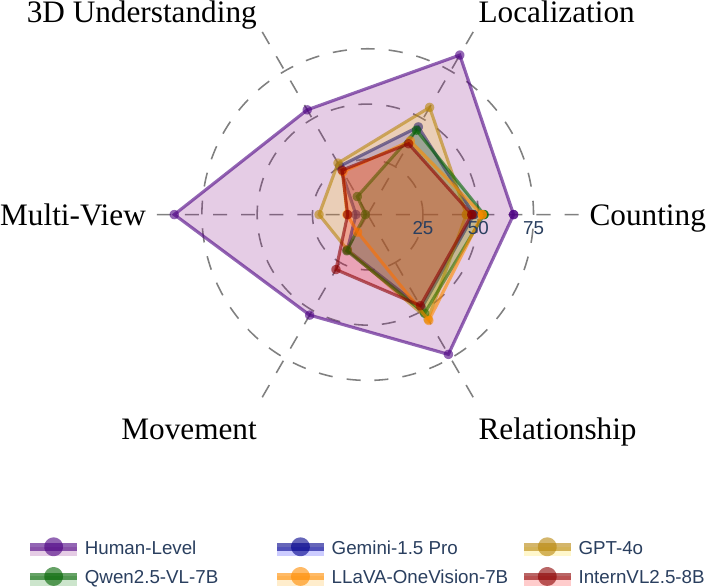}
    \caption*{(a) Different models' performance in a glance.}
  \end{minipage}
  \hfill
  \begin{minipage}[c]{0.48\textwidth}
    \centering
    \includegraphics[width=\linewidth]{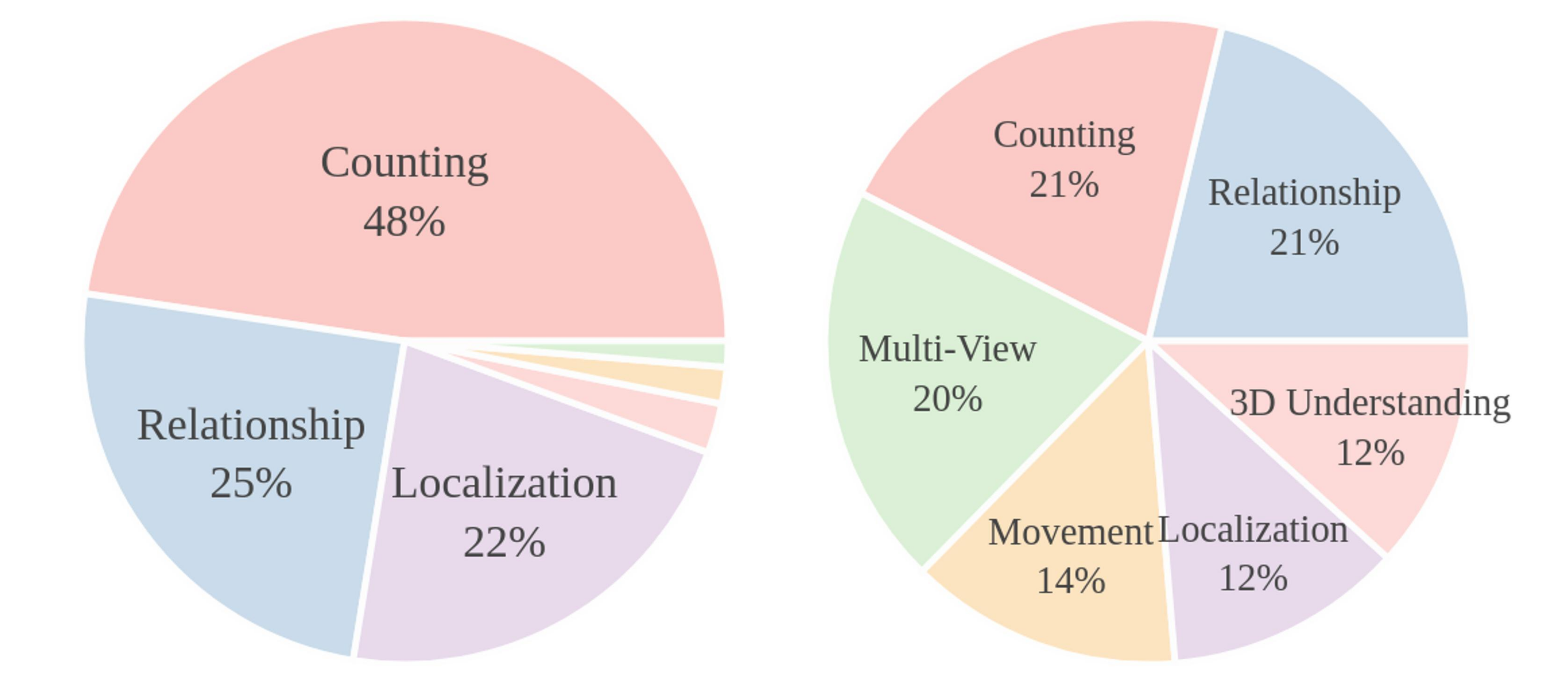}
    \vspace*{\fill}
    \caption*{(b) \textbf{Left:} spatial category distribution before balancing. \textbf{Right:} final benchmark's spatial category distribution.}
  \end{minipage}
  \caption{Data distribution and model's performance under six coarse spatial categories.}
  \label{fig:wide_layout}
\end{figure*}

\begin{figure*}
    \centering
    \includegraphics[width=0.95\linewidth]{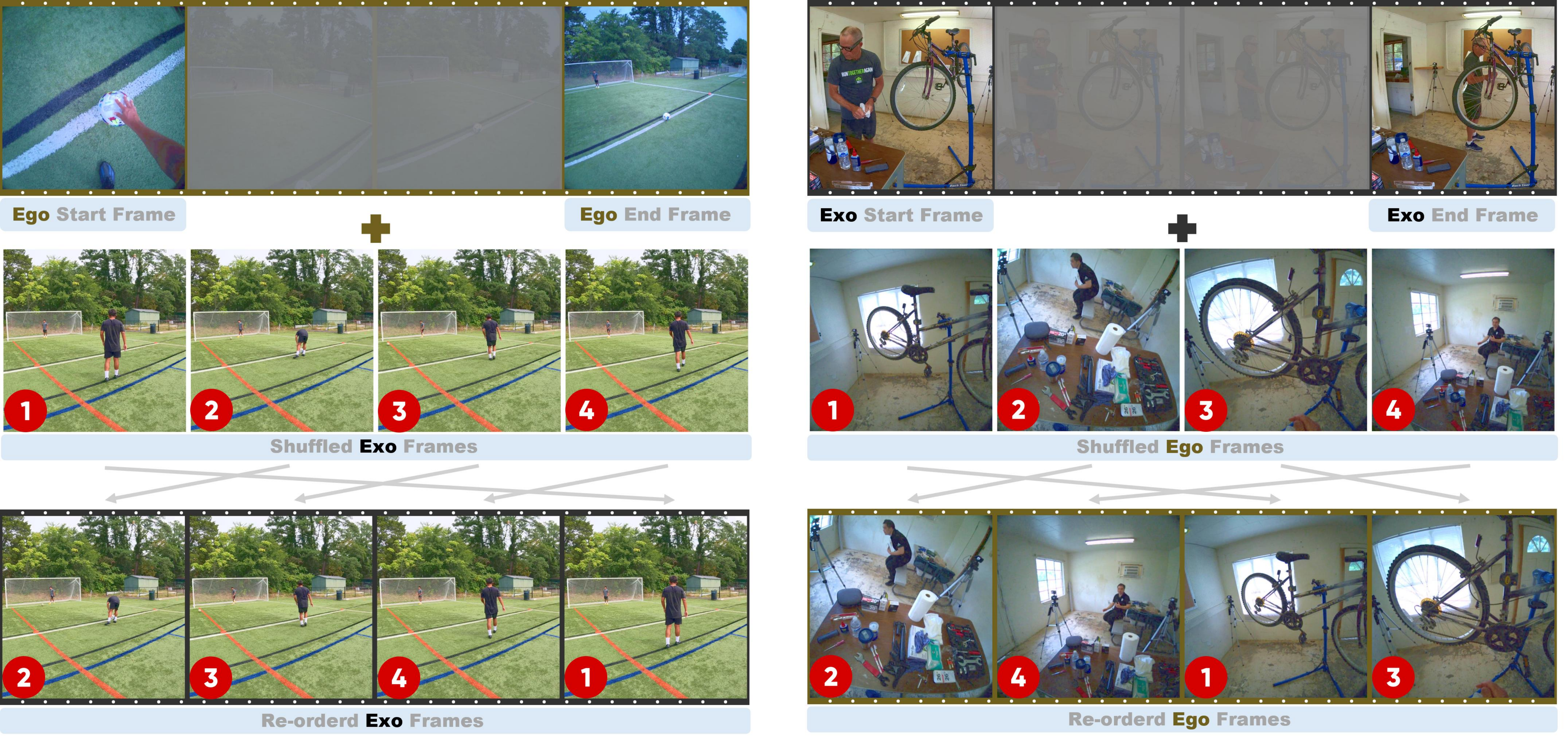}
    \caption{\textbf{Ego-Exo frames reordering tasks.} Given the start and end frames of a video clip in an egocentric view, and four randomly shuffled frames from the same clip in exocentric views, the model is tasked with reordering the four shuffled frames into their correct temporal sequence (or vice versa).}
        \label{fig:frames-reordering-tasks}
\end{figure*}

\begin{figure*}
\centering
\begin{minipage}{0.9\textwidth}
\begin{lstlisting}
enhanced_prompt_in_english = """
System Role Instructions (System):
You are ChatGPT, a large language model trained by OpenAI. Your goal is to help the user classify a series of Q&A pairs to determine whether they are spatially related. If a pair is indeed spatially related, you must further categorize it into one of the specified categories.

You must follow these rules:
1. If the Q&A content is NOT related to spatial relationships, simply answer:
   No
2. If the Q&A content IS related to spatial relationships, answer:
   Yes. This is a X problem because ...
   where X must be chosen from the following list:
   - Counting & Existence
   - Object Localization & Positioning
   - Spatial Relationship Reasoning
   - Depth & 3D Understanding
   - Multi-view & Cross-Image Reasoning
   - Movement Navigation & Intent Prediction
   - Other spatial category can not be sure.
3. If multiple Q&A pairs (N Q&A pairs) are provided in a single input, you must apply the same classification steps to each Q&A pair in the order they appear, and output the result for each pair in that order.

User Role Instructions (User):
Below are examples and their reference outputs (few-shot examples). Please study the logic and answer format shown in these examples before performing the classification:

[Example 1]
Input:
Question: How many blue floats are there?
Select from the following choices.
(A) 0
(B) 3
(C) 2
(D) 1
Answer: (D) 1

Output:
Yes. This is a Counting & Existence problem because it asks about the number of objects.

[Example 2]
Input:
Question: What is the position of the catcher relative to the home plate?
Options:
A: The catcher is to the left of the home plate.
B: The catcher is to the right of the home plate.
C: The catcher is behind the home plate.
D: The catcher is in front of the home plate.
Answer: A: The catcher is to the left of the home plate.

Output:
Yes. This is a Spatial Relationship Reasoning problem because it asks about the relative relation between two objects.


[Example 3]
Input:
Question: Where is the bongo?
Answer: On top of the brown shelf.

Output:
Yes. This is an Object Localization & Positioning problem because it asks about the location of an object.
"""

\end{lstlisting}   
\end{minipage}
\caption{Few-shot prompt used for spatial category classification. (Part I)}
\label{fig:prompt-1}
\end{figure*}

\begin{figure*}
\centering
\begin{minipage}{0.9\textwidth}
\begin{lstlisting}
enhanced_prompt_in_english += """
[Example 4]
Input:
Question: Here are some images and their corresponding depth images: <img><img><img><img>.
Please select the correct corresponding image for the target image: <img>.
The option images are: <img><img><img><img>
Answer: The second image.

Output:
Yes. This is a Depth & 3D Understanding problem because it asks about depth information.


[Example 5]
Input:
Question: These images are frames from a video. The video shows a static scene, and the camera is either moving clockwise (left) or counterclockwise (right) around the object.
The first image is from the beginning of the video, and the second image is from the end. Is the camera moving left or right during the filming?
Select from the following options:
(A) left
(B) right
Answer: (A) left

Output:
Yes. This is a Multi-view & Cross-Image Reasoning problem because it focuses on multi-view information of the object and determines the camera's rotation direction.


[Example 6]
Input:
Question: This is a navigation video of an agent following the instruction: "Exit the kitchen and wait in the sitting room, near the loveseat."
What is the next action it should take?
Options: Move forward. / Turn right and move forward. / Turn left and move forward. / Stop.
Answer: Stop

Output:
Yes. This is a Movement Navigation & Intent Prediction problem because it asks about the next action of the agent.


[Example 7]
Input:
Question: What is the color of the cat?
Answer: The cat is black.

Output:
No

[Example 8]
Input:
Question: Please correctly describe this set of images from a spatial context perspective.
Select from the following choices:
A: There is a box with four items, and three of them are touching the side.
B: There is a box with five items, all in the center.
C: There is a box with three items, and four of them are touching the side.
D: There is a bag with four items, and three of them are touching the side.
Answer: A.

Output:
Yes, but it's hard to determine the category. This is a spatially related problem because it asks about the spatial context of the objects.
(If you are unsure which exact category it belongs to, choose "Other spatial categories can not be sure.")
"""
\end{lstlisting}   
\end{minipage}
\caption{Few-shot prompt used for spatial category classification. (Part II)}
\label{fig:prompt-2}
\end{figure*}

\clearpage
\vspace{-20cm}
\begin{figure*}[t]
\centering
\begin{minipage}{0.9\textwidth}
\begin{lstlisting}
enhanced_prompt_in_english += """
[Main Task]
1. Read the new Q&A input(s).
2. First, decide whether each Q&A is related to spatial relationships.
3. If NOT related, simply output:
   No
4. If related, output:
   Yes. This is a [specific category] problem because [reason].
   where [specific category] is strictly from the list:
   - Counting & Existence
   - Object Localization & Positioning
   - Spatial Relationship Reasoning
   - Depth & 3D Understanding
   - Multi-view & Cross-Image Reasoning
   - Movement Navigation & Intent Prediction
   - Other spatial categories can not be sure.
5. If multiple Q&A pairs are given together (N Q&A pairs), repeat steps 2 to 4 for each Q&A pair in order, returning the results in the same order and prefixing each result with an index '1. ', '2. ', etc.

Please keep the output style consistent and follow all the rules above.
"""
\end{lstlisting}   
\end{minipage}
\caption{Few-shot prompt used for spatial category classification. (Part III)}
\label{fig:prompt-3}
\end{figure*}

\begin{figure*}
\centering
\includegraphics[width=0.8\linewidth]{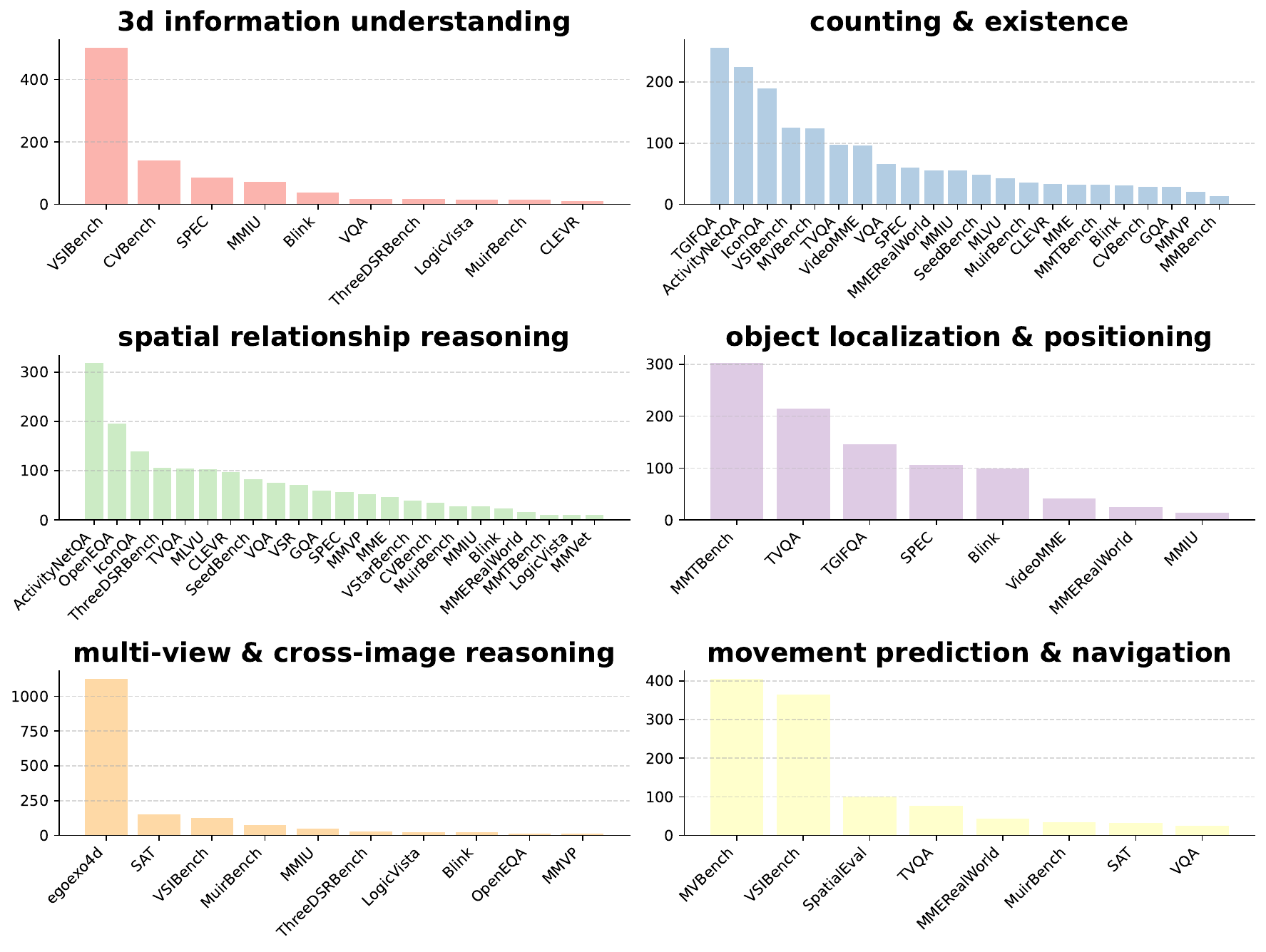}
\caption{Dataset composition for each spatial category.}
\label{fig:dataset-composition}
\end{figure*}


\end{document}